\documentclass{elsarticle}

\usepackage{hyperref}
\usepackage{amsmath,amsfonts,amssymb}
\usepackage{verbatim}
\usepackage{algpseudocode}
\usepackage{graphicx}
\usepackage{textcomp} %命令\textacutedbl的包,二阶导符号
\usepackage{multirow} %表格多行
\usepackage{epstopdf}
\usepackage{color}
\usepackage{algorithm}
\usepackage{threeparttable, booktabs}
\usepackage[table,xcdraw]{xcolor}
\usepackage{subfigure}
\usepackage{changepage}
\usepackage{bm}
\usepackage{diagbox}
\newtheorem{remark}{Remark}

\newcommand{\At}{\mathbf{A}}
\newcommand{\St}{\mathbf{S}}

\newcommand{\Yt}{\mathbf{Y}}
\newcommand{\Nt}{\mathbf{N}}

\newcommand{\Ft}{F}

\newcommand{\Pt}{\mathbf{P}}
\newcommand{\Qt}{\mathbf{Q}}
\newcommand{\Et}{\mathbf{E}}
\newcommand{\hEt}{\hat{\mathbf{E}}}

\newcommand{\mmSt}{\mathbf{\mathfrak{S}}}
\newcommand{\hmmSt}{\hat{\mathbf{\mathfrak{S}}}}
\newcommand{\Wt}{\mathbf{W}}

\newcommand{\gSt}{\ddot{\mathbf{S}}}%
\newcommand{\get}{\ddot{\mathbf{e}}}

\newcommand{\at}{\mathbf{a}}
\newcommand{\st}{\mathbf{s}}
\newcommand{\yt}{\mathbf{y}}
\newcommand{\nt}{\mathbf{n}}
\newcommand{\wt}{\mathbf{w}}
\newcommand{\xt}{\mathbf{x}}
\newcommand{\et}{\mathbf{e}}
\newcommand{\bt}{\mathbf{\beta}}
\newcommand{\mta}{\boldsymbol{\theta}}
\newcommand{\zta}{\boldsymbol{\zeta}}
\newcommand{\bmta}{\bar{\boldsymbol{\theta}}}

\newcommand{\ft}{\mathbf{f}}
\newcommand{\zt}{\mathbf{z}}

\newcommand{\bta}{\bar{\theta}}

\begin{document}
\begin{frontmatter}
	\title{Multiobjective Bilevel Evolutionary Approach for Off-Grid Direction-of-Arrival Estimation}
	\author[yb,ustc]{Bai Yan}
	\ead{yanb@sustech.edu.cn}
	
	\author[yb,ustc]{Qi Zhao}
	\ead{zhaoq@sustech.edu.cn}
	
	\author[yb]{Jin Zhang\corref{cor1}}
	\cortext[cor1]{Corresponding author}
	\ead{zhangj4@sustech.edu.cn}
	
	\author[az]{J. Andrew Zhang}
	\ead{Andrew.Zhang@uts.edu.au}
	
	\author[yb]{Xin Yao}
	\ead{xiny@sustech.edu.cn}	
	
	\address[yb]{Guangdong Provincial Key Laboratory of Brain-Inspired Intelligent Computation, Department of Computer Science and Engineering, Southern University of Science and Technology, Shenzhen 518055, China}
	
	\address[ustc]{School of Computer Science and Technology, University of Science and Technology of China, Hefei 230027, China}
	
	\address[az]{Global Big Data Technologies Centre, University of Technology Sydney, NSW 2007, Australia}
		
	\begin{abstract}
     The source number identification is an essential step in direction-of-arrival (DOA) estimation. Existing methods may provide a wrong source number due to inferior statistical properties (in low SNR or limited snapshots) or modeling errors (caused by relaxing sparse penalties), especially in impulsive noise. To address this issue, we propose a novel idea of simultaneous source number identification and DOA estimation. We formulate a multiobjective off-grid DOA estimation model to realize this idea, by which the source number can be automatically identified together with DOA estimation. In particular, the source number is properly exploited by the $l_0$ norm of impinging signals without relaxations, guaranteeing accuracy. Furthermore, we design a multiobjective bilevel evolutionary algorithm to solve the proposed model. The source number identification and sparse recovery are simultaneously optimized at the on-grid (lower) level. A forward search strategy is developed to further refine the grid at the off-grid (upper) level. This strategy does not need linear approximations and can eliminate the off-grid gap with low computational complexity. Simulation results demonstrate the outperformance of our method in terms of source number and root mean square error.
	\end{abstract}
		
	
\end{frontmatter}

\section{Introduction}
\begin{figure}[b]
	\centering
	\includegraphics[width=7cm,height=4.5cm]{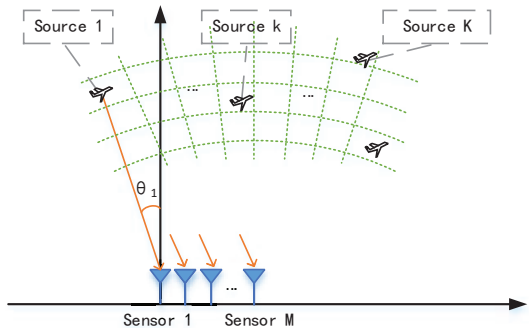}
	\caption{DOA estimation for location.}
	\label{fig-DOA}
\end{figure}

Direction-of-arrival (DOA) estimation refers to finding the direction information of electromagnetic waves/sources according to the outputs of receiving antennas/sensors that form an array, as shown in Fig. \ref{fig-DOA}. It is a crucial subject in array signal processing and is ubiquitous in radar, sonar, and wireless communications \cite{XiaJoint}. 

Since the number of sources is normally unknown a priori, methods for DOA estimation need to identify the source number as well as estimate the direction of each source. Existing methods can be divided into two categories. The first category is subspace-based methods \cite{RaoPerformance}, which exploits the eigenvalues decomposition of the sample covariance matrix of the array outputs. They use intuitive thresholding or information criterion to remove spurious sources from the obtained ones \cite{stoica2004model}\cite{valaee2004an}\cite{wax1985detection}, such that the source number is identified a posteriori.

The second category is sparse methods \cite{2016A}\cite{LIU2017171}\cite{YangOff}\cite{Carlin2013Directions}, which can be further divided into on-grid, gridless, and off-grid types. The on-grid type separates the continuous angular range into finite grid points and assumes sources fall exactly on the grid \cite{liu2017reweighted}, but this assumption is not always the case in practice. The gridless type directly operates in the continuous angular domain. This type needs to solve a time-consuming semidefinite programming problem and only works well in cases with well-separated sources \cite{wang2018ivdst:}\cite{wu2018a}. In contrast, the off-grid type adjusts the initial grid points to approach the sources \cite{YangOff}\cite{wu2018two}\cite{qi2018off}, which alleviates the gap between grid points and sources.

One major challenge in the above methods is the identification of source number. For subspace-based methods, the a posteriori processing tends to provide a wrong source number in limited snapshots or low signal-to-noise ratio (SNR) scenarios because of inferior statistical properties. For sparse methods, apart from postprocessing, $l_p$-norm ($0<p\leqslant1$) sparse penalties, Laplace prior \cite{YangOff}, and Gaussian scale mixtures \cite{Dai2017Sparse} have also been employed to capture the sparsity of impinging signals. These penalties/priors are relaxations of the $l_0$ norm of impinging signal matrix, which bring modeling error and bias the sparse distribution, subsequently overestimating the source number \cite{8963635}. 

To address the challenge in source number identification, in this paper, we propose a novel idea of simultaneous source number identification and DOA estimation. We design a multiobjective bilevel evolutionary approach (MoBEA) with two significant innovations to realize this idea. Firstly, we bulid a multiobjective DOA estimation model, in which the source number and a robust fitting error are taken as two conflicting objectives. Unlike existing DOA estimation models, our model can automatically identify the source number together with DOA estimation. Moreover, our model perfectly exploits the source number via the original $l_0$-norm of impinging signals without relaxation, providing a more precise source number. Besides, different from existing models explicitly or implicitly working with the assumption of Gaussian noise, we incorporate the robustness metric ``correntropy" \cite{Principe2010Information} into the model, greatly improving the performance in impulsive noise.

Secondly, we design a multiobjective bilevel evolutionary algorithm to solve the proposed model. Our algorithm includes on-grid (lower) level and off-grid (upper) level optimization. At the on-grid level, we develop a population-based genetic algorithm to simultaneously identify the source number and recover the impinging signal matrix on a coarse grid. The diversified solutions in the population evolve various source numbers and communicate with each other, providing multiple pathways to the optima. The knee solution is finally extracted from the obtained solutions, which represents the identified source number and active grid points. At the off-grid level, we introduce a forward search strategy to refine the active grid points. This strategy does not need linear approximations and can efficiently eliminate the off-grid gap with low computational complexity. Overall, MoBEA's main contributions are:
\begin{itemize}
	\item Multiobjective DOA estimation model. The proposed model can automatically identify the source number together with DOA estimation. Besides, it perfectly exploits the source number via the $l_0$-norm of impinging signals without relaxation, guaranteeing its accuracy. 
	\item Multiobjective bilevel evolutionary DOA estimation algorithm. It solves the proposed model via alternate execution of two levels' optimization. A forward search strategy is introduced in the off-grid level, which efficiently eliminates the off-grid gap with low computational complexity.
	\item Empirical validation of MoBEA's performance. Simulation results confirm the superiority of MoBEA in both source number identification and DOA estimation over state-of-the-art methods.
\end{itemize}

In this paper, bold-face letters represent vectors and matrices. $\mathbb{R}$ and $\mathbb{C}$ are the real and complex domain. $\mathbb{E}$ denotes the mathematical expectation. $T$, $*$, and $H$ denote transpose, conjugate, and conjugate transpose of a vector or matrix, respectively. $\At_{i,:}$ and $\At_{i,j}$ denote the $i$-th row and $(i,j)$-th element of matrix $\At$, respectively. $diag(\wt)$ is a diagonal matrix with $\wt$ as the diagonal elements. In particular, $\st|_\et$ stands for the sub-vector of $\st$ with entries indexed by the set $I=\{i|\et_i=1\}$. Similarly, $\St|_\et$ denotes the sub-matrix of $\St$ with rows indexed by the set $I=\{i|\et_i=1\}$. 

The rest of the paper includes related work in Section II, the sampling signal model in Section III, the proposed DOA estimation model and algorithm in Sections IV and V, respectively, simulation results in Section VI, and conclusions in Section VII.

\section{Related Work}
DOA estimation aims at finding the direction information of electromagnetic sources according to the outputs of receiving antennas/sensors that form an array. Methods for this purpose are mainly divided into subspace-based ones and sparse ones. Subspace-based methods exploit the eigenvalues decomposition of the sample covariance matrix of the array output \cite{RaoPerformance}. They employ intuitive thresholding or information criterion as postprocessing to determine the source number \cite{stoica2004model}\cite{valaee2004an}\cite{wax1985detection}. In scenarios with limited snapshots or low SNR, this postprocessing step may provide a wrong source number because of inferior statistical properties.

Sparse methods are inspired by the compressed sensing theory. Representatives include greedy algorithms \cite{2016A}, convex optimization approaches \cite{LIU2017171}, and sparse Bayesian learning (SBL)-based methods \cite{YangOff}\cite{Carlin2013Directions}. These methods show many prominent properties over subspace-based methods, e.g., enhanced robustness to lower SNR, fewer snapshots, and highly correlated sources \cite{chen2018sparse}\cite{hu2016source}. Sparse methods can be divided into on-grid, gridless, and off-grid types. The on-grid type separates the continuous angular range into finite grid points and assumes all sources fall exactly on the grid \cite{liu2017reweighted}. However, some sources may deviate from the initial grid in practice (i.e., the off-grid gap), degrading on-grid methods' performance.  

Gridless methods are based on the atomic norm theory \cite{wang2018ivdst:}\cite{wu2018a} or covariance fitting criterion \cite{yang2014a}. They operate in the continuous angular domain, thereby eliminating the off-grid gap. However, they need to solve a time-consuming semidefinite programming problem and only work well in cases with sufficiently separated sources. 

Off-grid methods are another type for alleviating the off-grid gap \cite{wu2018two}\cite{YangOff}\cite{qi2018off}. They parameterize the gap into the models, which modifies the initial grid to approach sources. Several strategies have been proposed for off-grid gap estimation, such as linear approximations \cite{wu2018two}\cite{YangOff}, orthogonality-based methods \cite{qi2018off}, and root-finding strategies \cite{Dai2017Sparse}\cite{zhang2019root}. Their performance depends on the trade-off between estimation accuracy and computational complexity. 

Apart from the aforementioned intuitive thresholding or information criterion-based methods \cite{stoica2004model}\cite{valaee2004an}\cite{wax1985detection}, $l_p$-norm ($0<p\leqslant1$) sparse penalties, Laplace prior \cite{YangOff}, and Gaussian scale mixtures \cite{Dai2017Sparse} have also been used in sparse methods to identify the source number. These penalties/priors are relaxations of the $l_0$ norm of impinging signal matrix, which bring modeling error and may not perfectly represent the sparse distributions, leading to overestimation of the source number \cite{8963635}. Identifying the source number is still an open challenge in DOA estimation.

In addition, DOA estimation methods are normally developed for Gaussian noise. However, the noise may exhibit non-Gaussian characteristics, e.g., impulsive noise (also known as burst noise) \cite{ZoubirRobust}, in practice. To deal with impulsive noise, maximum likelihood subspace methods \cite{kozick2000maximum-likelihood}\cite{zhang2017a} have been developed for Gaussian mixture model-based noise environments. Other robust subspace-based methods include fractional lower-order moment-based methods \cite{liu2001a}\cite{visuri2001subspace-based}, and the zero-memory nonlinear functions based methods \cite{swami2002on}. Different from them, the $l_p$-MUSIC considers the sample covariance matrix with a robust metric, i.e., minimizing the $l_p$-norm ($p\in[1,2)$) of the residual fitting error matrix to handle outliers \cite{Zeng2013}. In \cite{zhang2014a}\cite{wang2017a}, an effective correntropy induced estimator is integrated into MUSIC to reduce the influence of outliers. Several robust sparse methods have also been proposed. In \cite{TruongRobust}, an additional $l_2$-norm constraint on the parameter of interest is introduced to improve the estimation robustness. In \cite{wen2017robust} and \cite{shi2019robust}, robust relaxation methods by imposing a $l_p$ fidelity (with $p=1$ and $1.2$ respectively) for denoising are presented. Two robust SBL-based methods \cite{Dai2017Sparse} model the additive noise as the mix of dense Gaussian noise and sparse noise outliers to get rid of the adverse effect of noise outliers. However, the assumption that the impulsive noise has a sparse structure is not reasonable, leading to a performance loss \cite{dai2020robust}.

\section{Signal Model}
Consider $K$ narrowband far-field sources $\bar{s}_1(t)$, $\bar{s}_2(t)$, ..., $\bar{s}_K(t)$ impinging on a uniform linear array of $M$ omnidirectional sensors from directions of $\bmta=[\bta_1, \bta_2,..., \bta_K]^T$. The array output at time $t$ is modeled as %
\begin{equation}
\begin{aligned}
\yt(t)=\sum_{k=1}^{K}\at(\bta_k)\bar{s}_k(t)+\nt(t)=\At(\bmta)\bar{\st}(t)+\nt(t),
\end{aligned}
\label{eq-y=As+noise}
\end{equation}
where $\yt(t)=[y_1(t),y_2(t),...,y_M(t)]^T$ is the array output, $\bar{\st}(t)=[\bar{s}_1(t),\bar{s}_2(t),...,$ $\bar{s}_K(t)]^T$ are the signal waveforms, $\At(\bmta)=[\at(\bta_1),\at(\bta_2),...,\at(\bta_K)]$ is the array manifold matrix, $\at(\bta_k)=[1,e^{-j2\pi d/\lambda\sin(\bta_k)},...,e^{-j2\pi (M-1)d/\lambda\sin(\bta_k)}]^T$ contains the time delay of the $k$-th signal received at each sensor, $\lambda$ is the wavelength of sources, $d$ is the spacing distance between adjacent sensors, $\nt(t)=[n_1(t),n_2(t),...,n_M(t)]^T$ is an unknown noise vector. When $T$ snapshots are collected, the array output is 
\begin{equation}
\begin{aligned}
\Yt=\At(\bmta)\bar{\St}+\Nt,
\end{aligned}
\label{eq-Y=AS+noise}
\end{equation}
where $\bar{\St}=[\bar{\st}(1),\bar{\st}(2),...,\bar{\st}(T)]$, $\Yt=[\yt(1),\yt(2),...,\yt(T)]$, and $\Nt=[\nt(1),\nt(2),...,$ $\nt(T)]$. Given $\Yt$ and the mapping $\bmta\rightarrow\At(\bmta)$, the goal is to find DOAs with unknown $K$. 

\begin{figure}[t]
	\centering
	\includegraphics[width=7cm,height=3.8cm]{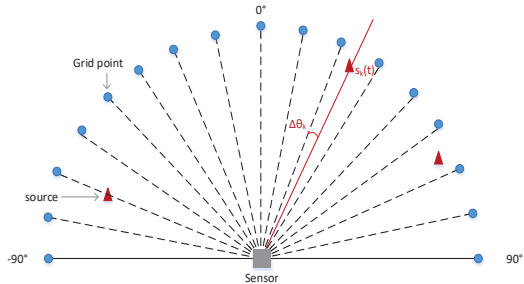}
	\caption{Off-grid case. The dotted lines stand for the predefined grid, and the red triangular symbol corresponds to a source. }
	\label{fig-signal model}
\end{figure}

To accomplish this goal, existing grid-based sparse methods cast DOA estimation as a sparse recovery problem \cite{YangOff}. The angular range $[-\pi/2, \pi/2]$ is divided into an equi-spaced grid $\mta_0=[\theta_1, ...,\theta_N]^T$ with the assumption that $N\gg K$, where $N$ denotes the total number of grid points, and $r=\theta_{i+1}-\theta_{i}$ is the grid interval. If all the sources fall exactly on this grid, it would be an \textit{on-grid} case. The array output $\Yt$ for the on-grid model is
\begin{equation}
\begin{aligned}
\Yt=\At(\mta_0)\St+\Nt,
\end{aligned}
\label{eq-Y=AS+noise(on-grid)}
\end{equation}
where the impinging signal matrix $\St$ is the extension of $\bar{\St}$ from $\bmta$ to $\mta_0$, with $\St$ being row-sparse, i.e., all columns of $\St$ are sparse and share the same support. However, some sources may deviate from the predefined grid (known as the \textit{off-grid} case), as shown in Fig. \ref{fig-signal model}. In this case, the sparse coefficient $\St$ is not sparse any more. Thus, the off-grid gap may lead to estimation degradation.

To alleviate the off-grid gap, the gap between true DOAs and the initial grid (denoted as grid mismatch $\zta$) is usually parameterized into the on-grid model (\ref{eq-Y=AS+noise(on-grid)}). As a result, the array output for the off-grid model can be formulated as
\begin{equation}
	\begin{aligned}
		\Yt=\At(\mta_0+\zta)\St+\Nt.
	\end{aligned}
	\label{eq-Y=AS+noise(off-grid)}
\end{equation}
The grid mismatch and impinging signal matrix are estimated by solving the following regularization optimization problem:  
\begin{equation} 
	\begin{aligned}
		\min_{\zta, \St}\ \rho\|\St\|_{2,0}+\frac{1}{2}\|\Yt-\At(\mta_0+\zta)\St\|^2_F, s.t.\ -\frac{r}{2}<\zta_i\leqslant\frac{r}{2},\ i=1,...N,
\end{aligned}
\tag{P1}\label{eq-Y-AS}
\end{equation}
where $\rho$ is a balancing parameter. $\|\Yt-\At(\mta_0+\zta)\St\|^2_F$ is a fitting error term, denoted as ``measurement error"; $\St$ is a row-sparse matrix whose $i$-th row corresponds to a possible source impinging on the array at $(\mta_0+\zta)_i$; $\|\St\|_{2,0}$ stands for the source number, which equals to the number of nonzero rows of $\St$: 
\begin{align}
	&\|\St\|_{2,0}=\|[\|\St_{1,:}\|_2, \|\St_{2,:}\|_2,..., \|\St_{N,:}\|_2]\|_0. \notag
\end{align}
$\|\St\|_{2,0}$ is a $l_0$-norm penalty. It is the proper and exact sparsity-enforcing penalty, which plays a prominent role in identifying the source number.   

\section{Proposed Multiobjective DOA Estimation Model}
Problem \eqref{eq-Y-AS} is NP-hard because of the $l_0$ norm\cite{Davis94adaptivenonlinear}. Existing off-grid methods \cite{2016A}\cite{YangOff}\cite{Carlin2013Directions} relaxed the $l_0$ norm to other suboptimal penalties. Such relaxation brings modeling error, which may provide some spurious DOAs and degrade the localization performance\cite{8963635}. Also, the performance is susceptible to the balancing parameter (i.e., $\rho$ in \eqref{eq-Y-AS}), which is hard to tune. 

To address the above issues, we naturally model the regularization optimization problem (\ref{eq-Y-AS}) as a multiobjective optimization problem (MOP). Our model holds two significant advantages: 1) it allows automatically identifying the source number together with DOA estimation, and 2) it perfectly exploits the source number via the $l_0$-norm of impinging signals without relaxation, guaranteeing accuracy. In detail, our model simultaneously minimizes the two conflicting objectives:
\begin{equation}
	\begin{aligned}
		\min_{\zta, \St}\ &(\|\St\|_{2,0}, \|\Yt-\At(\mta_0+\zta)\St\|_F^2), \\
		s.t.\ &-\frac{r}{2}<\zta_i\leqslant\frac{r}{2},\ i=1,...N,
	\end{aligned}
\tag{P2}\label{eq-MOEA(Y,AS)}
\end{equation}  
where $\|\St\|_{2,0}$ and $\|\Yt-\At(\mta_0+\zta)\St\|_F^2$ are the source number and measurement error, respectively. 

The MOP \eqref{eq-MOEA(Y,AS)} inherently involves two problems. 0ne is sparse recovery of the impinging signal matrix $\St$ with given grid, where $\|\St\|_{2,0}$ denotes the source number. The other is grid refinement with given $\St$. It should be noted that, compared to source number identification by sparse recovery, perceiving true DOAs is more critical. Therefore, the decision maker of the grid refinement problem should have the complete knowledge about sparse recovery, while the sparse recovery problem only observes the decisions of grid refinement. Therefore, we cast the MOP \eqref{eq-MOEA(Y,AS)} as a bilevel MOP:
\begin{equation}  
	\begin{aligned}  \min_{\zta, \St}\ &\|\Yt-\At(\mta_0+\zta)\St\|_F^2, \\
		s.t.\ &-\frac{r}{2}<\zta_i\leqslant\frac{r}{2},\ i=1,...N, \\
		&\St\in\arg\min_{\St}\ (\|\St\|_{2,0}, \|\Yt-\At(\mta_0+\zta)\St\|_F^2),
	\end{aligned}
	\tag{P3}\label{eq-MoBEA(Y,AS)}
\end{equation}
where the first and the third rows are the off-grid (upper) level and the on-grid (lower) level problems, respectively. The on-grid level simultaneously identifies the source number and recovers $\St$ with given grid. Based on the decisions of the on-grid level, the off-grid level minimizes the measurement error to refine the grid and update $\St$.  

The bilevel MOP \eqref{eq-MoBEA(Y,AS)} explicitly works in Gaussian noise. In practice, the impulsive noise may exist, bringing great challenges for accurate estimation. To cope with this, we incorporate a robust metric ``\textit{correntropy}" \cite{Principe2010Information} to reduce the detrimental effect of the impulsive noise. Correntropy is a local and nonlinear similarity measure in a feature space \cite{Principe2010Information}. Given two arbitrary vectors $\xt=[x_1,...,x_M]^T$ and $\zt=[z_1,...,z_M]^T$, the correntropy is approximated as \cite{Principe2010Information}
\begin{equation}
	\label{eq-correntropy}
	\begin{aligned}
		V(\xt,\zt):=&\text{E}[\kappa_\sigma(\xt-\zt)]\approx\frac{1}{M}\sum_{i=1}^{M}\kappa_\sigma(x_i-z_i),  \\
		\kappa_\sigma(\gamma)=&\frac{1}{\sqrt{2\pi}\sigma}\exp(-\frac{\gamma\gamma^*}{2\sigma^2}), 
	\end{aligned}
\end{equation}
where $\kappa_\sigma(\gamma)$ is the Gaussian kernel function with a kernel size $\sigma$. Based on correntropy, correntropy-based loss function ($\text{CLF}$) \cite{Wang2016Correntropy} is defined as
\begin{equation}
	\label{eq-CLF}
	\begin{aligned}
		V_{\text{CLF}}(\xt,\zt)=1-\frac{1}{M}\sum_{i=1}^{M}\exp(-\frac{(x_i-z_i)(x_i-z_i)^*}{2\sigma^2}).   
	\end{aligned}
\end{equation}
This function is related to the Welsch's cost function \cite{Principe2010Information}. Compared to the $l_2$-norm error term of the bilevel MOP \eqref{eq-MoBEA(Y,AS)}, $V_{\text{CLF}}$ increases much slower and is bounded, thus large noise outliers have a limited effect on $V_{\text{CLF}}$. Hence, it has better robustness to impulsive noise. With incorporating $V_{\text{CLF}}$, we give the final bilevel multiobjective optimization model
\begin{equation}  
	\begin{aligned} \min\ &\Ft(\St,\zta)=V_{\text{CLF}}(\Yt,\At(\mta_0+\zta)\St),   \\
		s.t.\ &-\frac{r}{2}<\zta_i\leqslant\frac{r}{2},\  i=1,...N, \\
		&\St\in\arg\min_{\St}\ft(\St)=(\|\St\|_{2,0}, V_{\text{CLF}}(\Yt,\At(\mta_0+\zta)\St)), 
	\end{aligned}
\tag{P4}\label{eq-MoBEA}
\end{equation}
where the first and the third rows are the off-grid level and on-grid level problems, respectively. The second row is the grid mismatch-related constraint. Compared to the regularization problem \eqref{eq-Y-AS}, no balancing parameter is needed. The sparsity of impinging signals is perfectly captured by the $l_0$ norm term without evoking relaxations. Hence the source number can be accurately identified. 

\begin{table}
	\caption{Notations in the proposed algorithm}
	\label{tab-descriptions}
	\renewcommand\arraystretch{1}
	\centering
	\resizebox{\textwidth}{!}{
	\begin{tabular}{ll}        
		\toprule
		Variable & Description \\ \midrule
		$G$ & the outer generation number \\ 
		$lt$ & the inner generation number at the on-grid level \\
		$ut$ & the inner generation number in forward search \\ 
		$\mta_0$	& the initial grid \\ 
		$\zta$  & the grid mismatch \\ 
		$\Et^G=\{\et_i^G\}_{i=1}^{\bar{N}}$ & the active sets at $G$-th generation  \\ 
		$\mmSt^G=\{\St_{i}^G\}_{i=1}^{\bar{N}}$ & the set of signal matrices decoded from $\Et^G$ \\      		
		$\Pt^G=(\Et^G,{\zta}^G)$ & the population at $G$-th generation \\ 
		$\ddot\et^G$ & the active set of the knee solution at $G$-th generation \\ 
		$\gSt^G$ & the impinging signal matrix of the knee solution at $G$-th generation \\ \bottomrule
	\end{tabular}} 
\end{table}
\section{Proposed Multiobjective BiLevel Evolutionary DOA Estimation Algorithm}
We design a multiobjective bilevel evolutionary algorithm to solve the proposed bilevel MOP \eqref{eq-MoBEA}. The designed algorithm has three key features: 1) it simultaneously identifies the source number and recovers the impinging signal matrix on a coarse grid at the on-grid level; 2) A forward search strategy is developed at the off-grid level, which eliminates the off-grid gap with lower computational complexity than other grid refinement strategies \cite{wu2018two}\cite{qi2018off}\cite{zhang2019root}; and 3) the population-based evolutionary algorithm evolves diversified search pathways to the optima, promoting the effectiveness and efficiency of the algorithm. 

\begin{figure}[t]
	\centering
	\includegraphics[width=6.8cm,height=8.8cm]{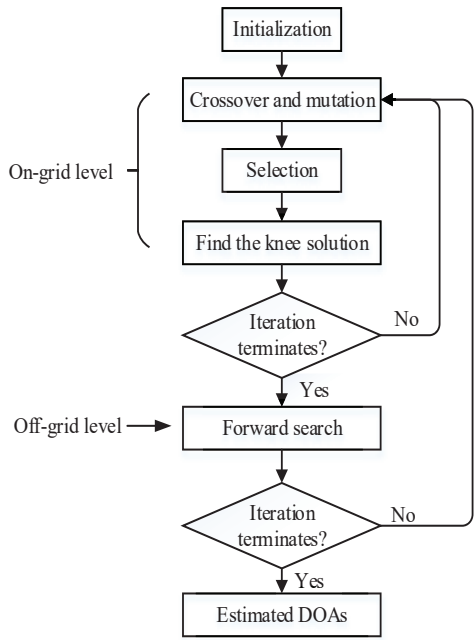}
	\caption{Framework of the proposed algorithm.}
	\label{fig-MoBEA}
\end{figure}

The workflow of the proposed algorithm is shown in Fig. \ref{fig-MoBEA}, with the pseudocode and variable definitions exhibited in Algorithm \ref{al-EBA} and Table \ref{tab-descriptions}, respectively. It starts with initialization, following by a bilevel optimization. At the on-grid level, an improved framework based on NSGA-II \cite{deb2002fast} is developed to simultaneously identify the source number and recover the impinging signal matrix $\St$. Instead of searching the complex-valued $\St$ directly, we encode $\St$ into a binary vector for simpleness. After evolutionary search and selection, the knee solution is extracted to represent the active grid points, where ``active'' means a source located at that grid point. After that, at the off-grid level, a forward search strategy is developed and executed for refining the grid represented by the knee solution. The population is then updated. When the stopping criterion is satisfied, the active grid points are the estimated DOAs. Key components of the proposed algorithm are detailed below.

\begin{algorithm}[t]
	\caption{Pseudocode of the proposed algorithm} 
	\begin{algorithmic}[1]
		\Require $\At$, $\Yt$, $\mta_0$, $\zta^0\leftarrow0^{N\times1}$  
		\Ensure	$\hat{\mta}$
		\State $G=0$;
		\State $(\Pt^G, \mmSt^G, \get^G, \gSt^G)=$\textit{Initialization}$(\At,\Yt, \zta^0)$; 	
		\While {``the algorithm does not teminate"}		
		\State $(\Pt^G, \mmSt^G, \get^G, \gSt^G)=$\textit{On\_grid}$(\Pt^G, \mmSt^G, \get^G,$ $\gSt^G)$;			
		\State $\zta^{G+1}=$\textit{Forward\_search}$(\Pt^G, \mmSt^G,$ $\get^G, \gSt^G)$, where $\Pt^G=(\Et^G,{\zta}^G)$;
		\State /*Population updates*/	
		\State $\Pt^{G+1}\leftarrow(\Et^G, \zta^{G+1})$;	
		\State $\mmSt^{G+1}=$\textit{Decoding}$(\Et^G, \gSt^G, \zta^{G+1}, \sigma^G)$;
		\State $G=G+1$;	
		\EndWhile	
		\State $\hat{\mta}=(\mta_0+\zta^G)|_{\get^G}$.
	\end{algorithmic}
	\label{al-EBA}
\end{algorithm}

\subsection{Chromosome Encoding and Decoding}\label{sec-encoding}
$(\St,{\zta})$ represents the solution of the proposed model \eqref{eq-MoBEA}, but it is non-trivial to execute genetic operators on the complex-valued impinging signal matrix $\St$. Therefore, we search the locations of the nonzero rows of $\St$ instead and then recover these nonzero values. In detail, we encode $(\St,{\zta})$ as $(\et,{\zta})$, where $\et$ is a binary vector with ``1" and ``0" denoting the corresponding row of $\St$ being nonzero and zero entries, respectively. For unity, we denote $\et$ as ``active set''. 

To recover $\St$ from $\et$, we develop a decoding method improved from the correntropy matching pursuit (CMP) algorithm \cite{Wang2016Correntropy}. For clarity, we briefly introduce CMP before describing the decoding process. 

\subsubsection{Correntropy matching pursuit}
CMP is a matching pursuit method for sparse recovery in impulsive noise \cite{Wang2016Correntropy}. In compressive sensing, let the measurement being $\yt=\At\st+\nt$ with the assumption that the dictionary $\At$ and measurements $\yt$ are both known, CMP solves a correntropy loss minimization problem to recover $\st$
\begin{align}
\label{eq-CMP-prob}
\st&=\arg\min_{\st} V_{\text{CLF}}(\yt,\At\st)=1-\frac{1}{M}\sum_{i=1}^{M}\exp(-\frac{z_iz_i^*}{2\sigma^2}),                                      
\end{align}
where $\zt=\yt-\At\st$ is the residual. With the half-quadratic theory \cite{rockafellar1970convex}, problem (\ref{eq-CMP-prob}) can be reformulated as a weighted least square problem and solved by \cite{Wang2016Correntropy}:
\begin{align}
\label{eq-sigma-res}
\sigma^{l+1}&=(\frac{1}{2M}\|\yt-\At\st^l\|^2_2)^{\frac{1}{2}},\\
\label{eq-CMP-w}
w^{l+1}_i&=g_\sigma(y_i-A_{i,:}\st^l),\ i=1,2,...M,\\   
\label{eq-CMP-WLS} 
\st^{l+1}|_\et&=\mathop{\arg\min}_{\text{supp}(\st)\in\{i|\et_i=1\}}\|\sqrt{diag(\wt^{l+1})}(\yt-\At\st)\|^2_2 \notag\\
&=(\At^T diag(\wt^{l+1})\At)^{-1}\At^T diag(\wt^{l+1})\yt,       
\end{align}
where $\st^{l+1}|_{\et}$ is the sub-vector of $\st^{l+1}$ with entries indexed by $I=\{i|\et_i=1\}$, $g_\sigma(p)=\exp(-\frac{pp^*}{2\sigma^2})$, and $\text{supp}(\st)$ denotes the nonzero entries of $\st$. $\wt$ is a weight vector. It indicates the importance of measurements, i.e., the small coefficients in $\wt$ suppress the severely contaminated measurements, and the large coefficients enable to preserve the clean measurements. $\wt$ can improve CMP to identify the true atoms, thus boosting the recovery performance \cite{Wang2016Correntropy}.

\subsubsection{Decoding}  
We extend the CMP to the case of multiple measurement vectors, as the decoding process of the proposed algorithm (see Algorithm \ref{al-decode}). In Algorithm \ref{al-decode}, $\Wt$ is a weight matrix with each column being the weight vector at each snapshot. To decode the active set, we first compute the weight matrix based on the impinging signal matrix of the knee solution $\gSt$, where the knee solution will be detailed in Section \ref{sec-initialization}. Then, the impinging signal matrix is recovered by solving a series of reweighted least square problems.  

In the decoding process, the choice of kernel size $\sigma$ is important. To balance the recovery accuracy and convergence speed, we employ the kernel annealing method \cite{Principe2010Information} instead of (\ref{eq-sigma-res}) due to its greater performance enhancement found in \cite{He2018Maximum}. Specifically, the kernel size is defined by 
\begin{equation} 
\label{eq-sigma}
\begin{aligned}
\sigma(G)=\sigma_{max}\exp(-\nu G)+\sigma_{min},
\end{aligned}
\end{equation}  
where $G$ is the generation counter of the outer cycle, $\sigma_{min}$=0.03, $\sigma_{max}$ is calculated by $0.5(abs(\Yt)_{0.875}-abs(\Yt)_{0.125})-\sigma_{min}$ with $abs(\Yt)_l$ being the $l$-th quantile of $\{|Y_{i,j}|\}^{M\times T}$, and $\nu=2\times10^{-4}$ is the decay rate. 

\begin{algorithm}[t]
	\caption{\textit{Decoding}} 
	\begin{algorithmic}[1]
		\Require $\At$, $\Yt$, $\Et$, $\gSt$, ${\zta}$, $\sigma$, $\mmSt=\varnothing$ 
		\Ensure	$\mmSt$
		\State $W_{i,j}=g_\sigma((\Yt-\At\gSt)_{i,j})$, where $g_\sigma(p)=\exp(-\frac{pp^*}{2\sigma^2})$,\ $i=1,...,M$,\ $j=1,...,T$; 
		\For{$\text{each active set}\ \et\in\Et$}
		\State $\St\leftarrow0^{N\times T}$;
		\For{$j=1:T$}
		\State $\St_{:,j}|_\et=(\At^T diag(\Wt_{:,j})\At)^{-1}\At^T diag(\Wt_{:,j})\Yt_{:,j}$; 	
		\EndFor	
		\State $\text{put}\ \St\ \text{into}\ \mmSt$;
		\EndFor
	\end{algorithmic}
	\label{al-decode}
\end{algorithm}

\subsection{Initialization} \label{sec-initialization}
We leverage the statistical correlation knowledge to provide a preliminary choice for the active set to enhance the search efficiency. Since a uniform linear array can identify up to $M-1$ sources, the condition $K\leq M-1$ can be considered as a priori \cite{ottersten1998covariance}. In noise-free cases, the sources can be roughly located by selecting the $K$ atoms that are most correlated with the measurements. Motivated by this and consider the atoms' rank distortion caused by impulsive noise, we give priority to the atoms that have larger correlations, e.g., those who rank the top $2M$. In this way, some proper atoms are involved with great probability, thus improving the search efficiency. The process is shown in Line 1-5 of Algorithm \ref{al-initialization}. For the missing atoms, they can be further searched in the mutation step of the on-grid level.

\begin{algorithm}[t]
	\caption{\textit{Initialization}}
	\begin{algorithmic}[1] 
		\Require $\At$, $\Yt$, ${\zta}^0$
		\Ensure $\Pt^0$, $\mmSt^0$, $\get^{0}$, $\gSt^{0}$
		\State $\mathbf{\tau}=\sum_{t=1}^{T}|<\yt(t), A_{:,j}>|$, $j=1,...,N$;	
		\State $\mathbf{\Lambda}\leftarrow$ save the indices of $2M$ highest correlation in $\mathbf{\tau}$;
		\State $\Et^0\leftarrow$ randomly choose no more than $M-1$ indices from $\mathbf{\Lambda}$ to construct $\bar{N}$ active sets;
		\State $\Pt^0=(\Et^0, {\zta}^0)$;	\label{al-initialization-combination}
		\State $\mmSt^0=$\textit{Decoding}$(\Et^0, 0^{N\times T}, {\zta}^0, \sigma)$;\ // Algorithm \ref{al-decode}	
		\State $\get^0, \gSt^{0} \leftarrow$\textit{Knee\_Identification}$(\Pt^0$, $\mmSt^0$).  \label{al-initialization-find knee}			
	\end{algorithmic}
	\label{al-initialization}
\end{algorithm}

After that, the knee solution is selected from the current population for preparing the next decoding. The reason for the above operation is that in each decoding process, a solution is required to estimate the weight matrix. Here we choose the knee solution based on the following considerations. The knee solution is commonly an interesting point along the Pareto front (PF), and it provides a promising trade-off between the two objectives \cite{RachmawatiMultiobjective}. Therefore, it is expected to use the knee solution to obtain reasonable weights and bring performance enhancement. Even when the knee solution fails to be the most satisfactory, it is still a Pareto nondominate solution from the perspective of multiobjective optimization. Here the kink method \cite{mierswa2006information} is employed to identify the knee solution, where the solution with the largest slope variance over the PF is the knee solution. 

\subsection{On-grid Level Optimization}
At the on-grid level, we develop a framework based on NSGA-II \cite{deb2002fast} to simultaneously identify the source number and recover the signals, where the grid mismatch acts as parameters. The procedure is exhibited in Algorithm \ref{al-on-grid}. The active set $\Et^{lt}$ firstly undergoes the one-point crossover, and the bitwise mutation operations \cite{PoliSchema} to create offspring solution set $\hEt^{lt}$. $\hEt^{lt}$ is then decoded according to Algorithm \ref{al-decode} to recover the corresponding signals $\mmSt^{lt}$. After that, the environmental selection operator of NSGA-II \cite{deb2002fast} is executed to select $\bar{N}$ elite solutions for the next generation. Lastly, the knee solution is identified. This solution will be used to estimate the weight matrix in the next decoding and to identify the active grid points for off-grid level refinement. 

\begin{algorithm}[t]
	\caption{\textit{On-grid}}
	\begin{algorithmic}[1]	
		\Require $\Pt^{lt}=[\Et^{lt}, {\zta}]$, $\mmSt^{lt}$, $\get^{lt}$, $\gSt^{lt}$
		\Ensure $\Pt^{lt}$, $\mmSt^{lt}$, $\get^{lt}$, $\gSt^{lt}$
		\For{``$lt=1:lt_{max}$"}		
		\State $\hEt^{lt}=$\textit{Crossover\&Mutation}$(\Et^{lt}$); \label{step-recombination} 
		\State $\hmmSt^{lt}=$\textit{Decoding}$(\hEt^{lt}, \gSt^{lt}, {\zta}, \sigma)$;\ // Algorithm \ref{al-decode}	
		\State $\Qt^{lt}=[\hEt^{lt}, {\zta}]$;	
		\State $\Pt^{lt+1}, \mmSt^{lt+1}\leftarrow$\textit{Environmental\_Selection}$(\Pt^{lt}\cup\Qt^{lt}, \mmSt^{lt}\cup\hmmSt^{lt})$\nonumber; \label{step-selection}    
		\State $\get^{lt+1}, \gSt^{lt+1}\leftarrow$\textit{Knee\_Identification}$(\Pt^{lt+1}$, $\mmSt^{lt+1}$);			
		\EndFor
	\end{algorithmic}
	\label{al-on-grid}
\end{algorithm}

\subsection{Forward Search}
We propose a straightforward forward search strategy at the off-grid level for solving the grid mismatch. Unlike existing grid refinement strategies, this strategy does not require modeling approximation, hence achieving higher DOA estimation quality. To save computational cost, the grid points corresponding to the knee solution's nonzero entries are viewed as ``active" and allowed to be refined. The perturbation direction of each active grid point where to decrease the correntropy-based loss function is obtained separately. Then all the active grid points are perturbed together along with the obtained directions.

The procedure of the forward search is shown in Algorithm \ref{al-off-grid}. The indices of active grid points are saved into $I$ according to the active set of the knee solution $\get$. Subsequently, the perturbation directions of the active grid points are detected and saved into $\mathbf{\beta}$ (Lines 3 to 10 of Algorithm \ref{al-off-grid}): for each active grid point, a random perturbation direction is given, and it is accepted only if the corresponding correntropy-based loss function decreases; otherwise, the perturbation direction is switched to the opposite direction or zero. After that, the active grid points are perturbed together along with $\mathbf{\beta}$ with a stepsize $\mu$ (Line 13), where $\mu$ is set to $r/100$ to provide a high resolution. The perturbation is repeated several times until $ut$ reaches its maximum value $ut_{max}$ or the correntropy-based loss function no longer decreases.             

\begin{algorithm}[t]
	\caption{\textit{Forward search}}
	\begin{algorithmic}[1]
		\Require $\Pt=[\Et, {\zta}]$, $\mmSt$, $\get$, $\gSt$
		\Ensure ${\zta}$ 
		\State Find the indices of active grid points: $I=\{i|\get_i=1\}$;
		\For {each index $i\in I$,}
		\State give a random perturbation direction: $\beta_i=-1\ or\ +1$;
		\State $\hat{\zta_i}=\zta_i+\mu \beta_i$;
		\If {$\Ft(\gSt, \hat{{\zta}})>\Ft(\gSt, {\zta})$}
		\State $\beta_i=-\beta_i$;
		\ElsIf{$\Ft(\gSt, \hat{{\zta}})=\Ft(\gSt, {\zta})$}
		\State $\beta_i=0$;
		\EndIf
        \EndFor 
        \State ${\zta}^1={\zta}$;
		\For{``$ut=1:ut_{max}$"}			
        \State ${\zta}^{ut+1}={\zta}^{ut}+\mu\bt$;	 
        \If {$\Ft(\gSt, {\zta}^{ut+1})\geqslant\Ft(\gSt, {\zta}^{ut})$}
        \State $\zta\leftarrow {\zta}^{ut}$
        \State \text{break};
        \EndIf
		\EndFor
	\end{algorithmic}
	\label{al-off-grid}
\end{algorithm}

\begin{remark}
	For any active grid point, it is finally accepted if it is within the set of $-r/2<\zta_i\leqslant r/2$. Otherwise, it is rejected, and the corresponding grid point remains unchanged.
\end{remark}

\subsection{Computational Complexity}\label{sec-complexity}
The main computational complexity of the proposed algorithm lies in the decoding process, i.e., Line 5 of algorithm \ref{al-initialization}, Line 3 of Algorithm \ref{al-on-grid} and Line 22 of Algorithm \ref{al-off-grid}. For each solution, the complexity of decoding is $T\times O(N^3)$ in the worst case, where $T$ and $N$ are the number of snapshots and the number of grid points, respectively. Thus, the algorithm's complexity is $(lt+1)\bar{N}T\times O(N^3)$, where $lt$ is the number of inner iterations of the on-grid level, $\bar{N}$ is the population size. It can be seen that the number of grid points has a significant impact on the computational complexity. To reduce the execution time, the decoding is suggested to be implemented in parallel in practical applications.      

\section{Simulation Results}\label{sec-simulation}
In this section, we first conduct simulations to investigate the effectiveness of operators in the proposed MoBEA. Then, we compare MoBEA with the state-of-the-art robust algorithms, i.e., $l_p$-MUSIC \cite{Zeng2013}, MCC-MUSIC \cite{wang2017a}, Bayes-optimal \cite{Dai2017Sparse}, and Fast-alternating \cite{Dai2017Sparse} in terms of root mean square error and average identified source number. The comparison algorithms are introduced below.
\begin{itemize}
	\item $l_p$-MUSIC: A representative robust subspace-based algorithm that adopts the $l_p$-norm of the residual fitting error matrix for subspace decomposition.
	\item MCC-MUSIC: A state-of-the-art robust subspace-based algorithm that estimates the signal subspace by solving an optimization problem under the maximum correntropy criterion.
	\item Bayes-optimal: A state-of-the-art off-grid SBL-based algorithm that models the measurement noise as the mix of Gaussian noise and outliers.	
	\item Fast-alternating: A state-of-the-art off-grid SBL-based algorithm which is a fast execution version of the Bayes-optimal algorithm.
\end{itemize} 

\subsection{Simulation Settings}
The comparison algorithms need the source number $K$ as the input. Here we set $K=M-1$ \footnote{A uniform linear array can identify up to $M-1$ sources, the condition $K\leq M-1$ can be considered as a priori while the exact value of $K$ is unavailable \cite{ottersten1998covariance}.}. For $l_p$-MUSIC, we set $p=1.1$. The grid interval is set to $0.1^\circ$ for $l_p$-MUSIC and MCC-MUSIC, and $2^\circ$ for the remaining methods if not stated. For MoBEA, the crossover probability and the mutation probability are set to 0.9 and $1/N$, respectively, following the practice of \cite{deb2002fast}. The population size $\overline{N}$ is 50. The on-grid level optimization terminates when its inner generation reaches 50 or the knee solution remains unchanged in five consecutive generations. 

In all simulations, we consider a small number of uncorrelated sources impinging on a uniform linear array of sensors with an inter-sensor spacing of $d=\lambda/2$. The number of receiving sensors is $M=8$. For a fair comparison, all methods stop running when the variance change of the impinging signal $\St^G$ is less than $10^{-6}$ in five consecutive generations, or the outer-cycle generation reaches 200; and the total number of Monte Carlo trials is 100 for all algorithms. All the experiments are implemented in MATLAB R2018b on a laptop with Intel i5-8265U CPU and 8GB RAM.

Two widely-used PDF models, i.e., Gaussian mixture model (GMM) \cite{kozick2000maximum-likelihood}, and symmetric $\alpha$-stable distribution (S$\alpha$S) \cite{1995Signal}, are considered to model the impulsive noise. All sources are assumed to possess the equal power $\eta_s^2$. 

\subsubsection{GMM}
The PDF of a two-term Gaussian mixture noise $\nt(t)$ can be described as
\begin{equation} 
	\label{eq-GMM}
	\begin{aligned}
		p_{\nt}(x)=\sum_{i=1}^{2}\frac{c_i}{\pi\eta_i^2}\exp(-\frac{|x|^2}{\eta_i^2}),
	\end{aligned}
\end{equation}
where $0\leq c_i\leq1$ and $\eta_i^2$ are the probability and variance of the $i$-th term with $c_1+c_2=1$, respectively. Assume $\eta_2^2=100\eta_1^2$ and $c1>c2>0$, the $\text{SNR}$ noise model can be viewed as the large noise outliers of variance $\eta_2^2$ with a smaller probability $c_2$  embedded into Gaussian noise of variance $\eta_1^2$ with a larger probability $c_1$. According to \cite{Dai2017Sparse}, the $\text{SNR}$ is simplified as $\text{SNR}=\eta_s^2/\eta_1^2$.

\subsubsection{$\alpha$-Stable distribution}
The symmetric S$\alpha$S distribution with zero-location is used to model the noise. Its characteristic function is defined as
\begin{equation} 
	\label{eq-sas}
	\begin{aligned}
		\varphi(x)=\exp(-\gamma^\alpha|x|^\alpha),
	\end{aligned}
\end{equation} 
where $0<\alpha\leq2$ denotes the characteristic exponent that indicates the tail of the distribution, and $\gamma$ is the scale. When $\alpha=2$ and $\alpha=1$, the S$\alpha$S reduces to the Gaussian distribution and the Cauchy distribution, respectively. The smaller value of $\alpha$, the more impulsive noise is. Since the closed-form PDF of S$\alpha$S does not exist when $\alpha\neq2$ and $\alpha\neq1$ \cite{1995Signal}, the $\text{SNR}$ becomes meaningless. Instead, the generalized $\text{SNR}$ ($\text{GSNR}$) \cite{Zeng2013} is reformulated as $\text{GSNR}=\eta_s^2/\gamma^\alpha$. 

\subsection{Performance metrics}
Since the Bayesian-based algorithms and the proposed MoBEA do not output the spatial spectrum, the spatial spectrum will not be used for comparison in this paper. We employ two statistical measures, i.e., the root mean square error ($\text{$\text{RMSE}$}$), and the average estimated source number.

Let $(\tilde{\theta}_k)_i$ stands for the estimate of $\bta_k$ at the $i$-th Monte Carlo trial, the $\text{RMSE}$ can be formulated as
\begin{equation} 
\label{eq-RMSE}
\begin{aligned}
\text{RMSE}=\sqrt{\frac{1}{K\upsilon}\sum_{i=1}^{\upsilon}\sum_{k=1}^{K}((\tilde{\theta}_k)_i-\bta_k)^2},
\end{aligned}
\end{equation} 
where $K$ is the true source number, and $\upsilon=100$ is the total number of Monte Carlo trials. For a given simulation point, the RMSE is obtained by averaging only the trials in which the estimated source number is greater than or equals to $K$. The assignment of estimated DOAs to the true one is executed based on the Hungarian algorithm \cite{1962Algorithms}. The average estimated source number is obtained by averaging the empirical source numbers of all Monte Carlo runs. These two metrics well reveal the identification capacity of the DOAs' positions and number.

\subsection{Detailed Analysis of MoBEA} \label{sec-exp-MoBEA}
In this subsection, the efficacy of the knee solution and the forward search scheme are investigated to demonstrate the superior performance of MoBEA.

\subsubsection{Efficacy of the knee solution} \label{sec-importance of knee}
At the on-grid level, the knee solution extracted from the PF is employed to represent the active grid points for grid refinement. It provides a promising trade-off between the source number and the estimation accuracy. 

\begin{figure*}[] 
	\centering
	\subfigure[]{\includegraphics[width=0.32\textwidth]{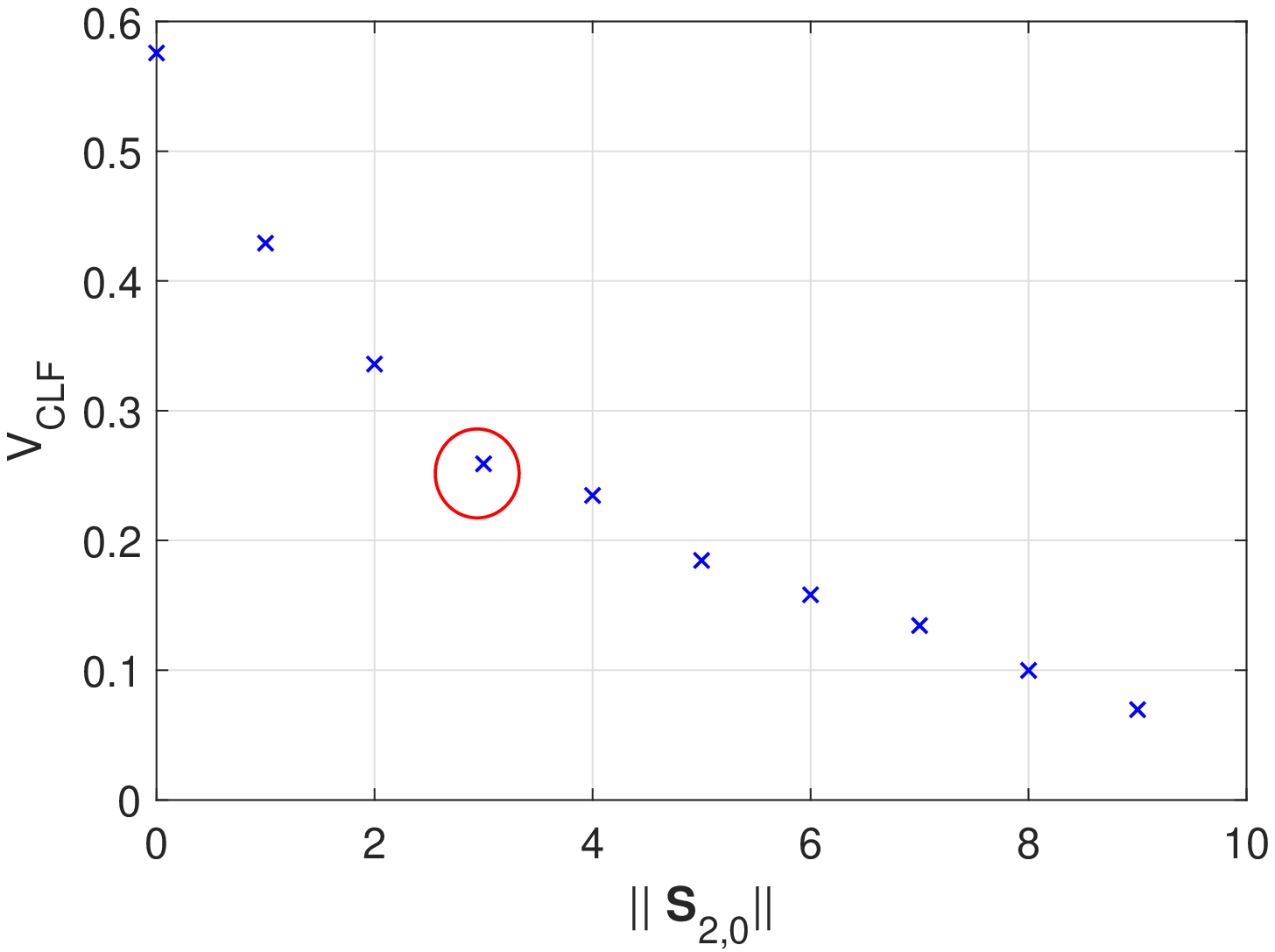}}
	\subfigure[]{\includegraphics[width=0.32\textwidth]{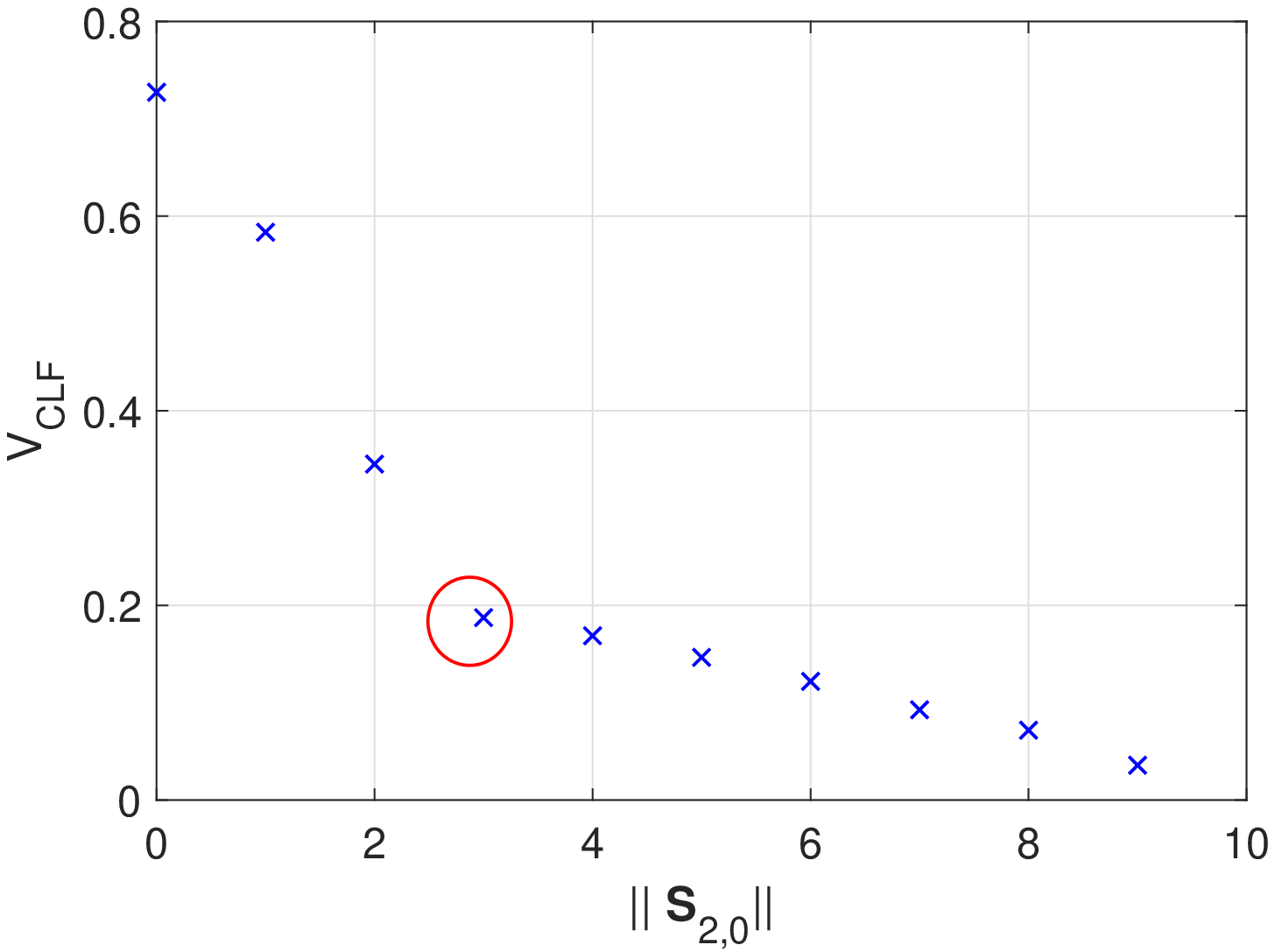}}
	\subfigure[]{\includegraphics[width=0.32\textwidth]{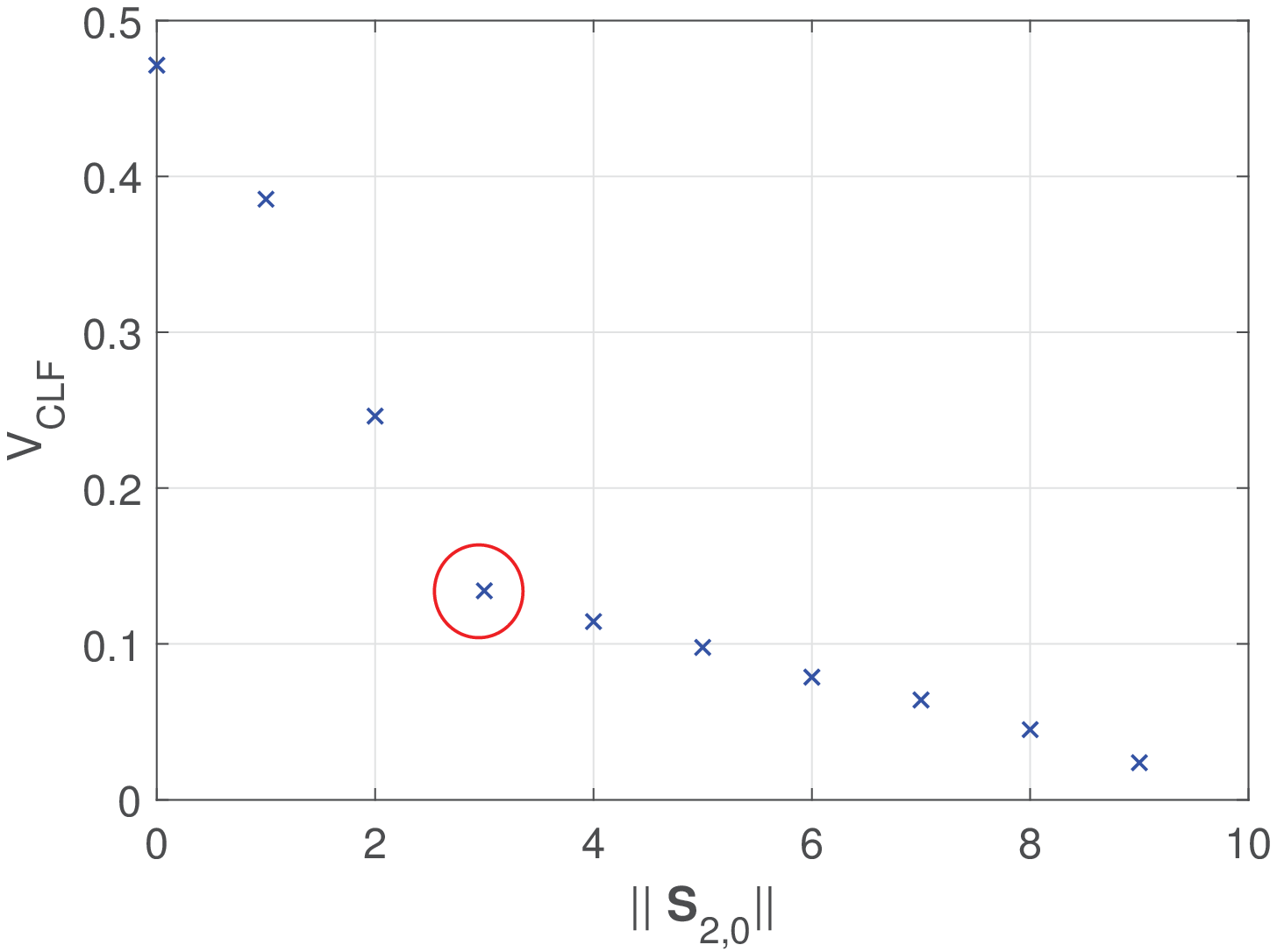}}
	\caption{The final PF obtained at the on-grid level in S$\alpha$S noise ($\alpha=1.4$). (a) GSNR=0dB. (b) GSNR=5dB. (c) GSNR=10dB. The solution in the red circle is the identified knee solution.}
	\label{fig-PF}
\end{figure*}

We first investigate the relationship between the knee solution and the source number. Fig. \ref{fig-PF} shows the typical final PF obtained by the on-grid level optimization of 100 trails under different GSNRs in S$\alpha$S noise with $\alpha=1.4$. The corresponding settings are as follows: three uncorrelated sources are from $-9.7^{\circ}$, $6.8^{\circ}$, and $12.7^{\circ}$. The number of snapshots is $20$. Since a uniform linear array can identify up to $M-1$ sources \cite{ottersten1998covariance}, it is expected to identify the knee solution from the non-dominated solutions whose source number is less than $M$ (where $M=8$). It can be observed that the knee solution in the red circle corresponds to the true source number $K=3$. 

\begin{figure}[] 
	\centering
	\subfigure[RMSE]{\includegraphics[width=0.45\textwidth]{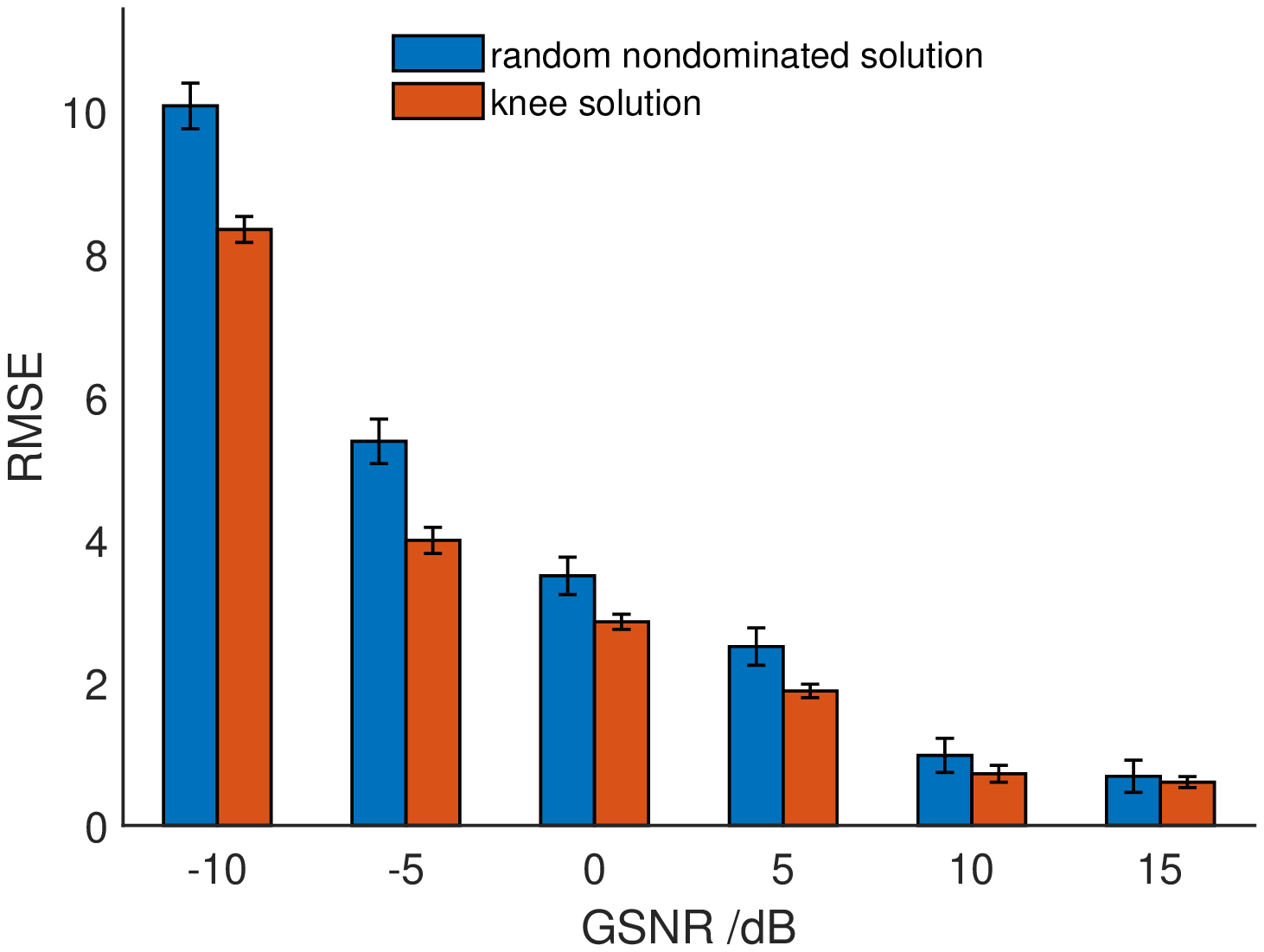}}
	\subfigure[Average estimated source number]{\includegraphics[width=0.45\textwidth]{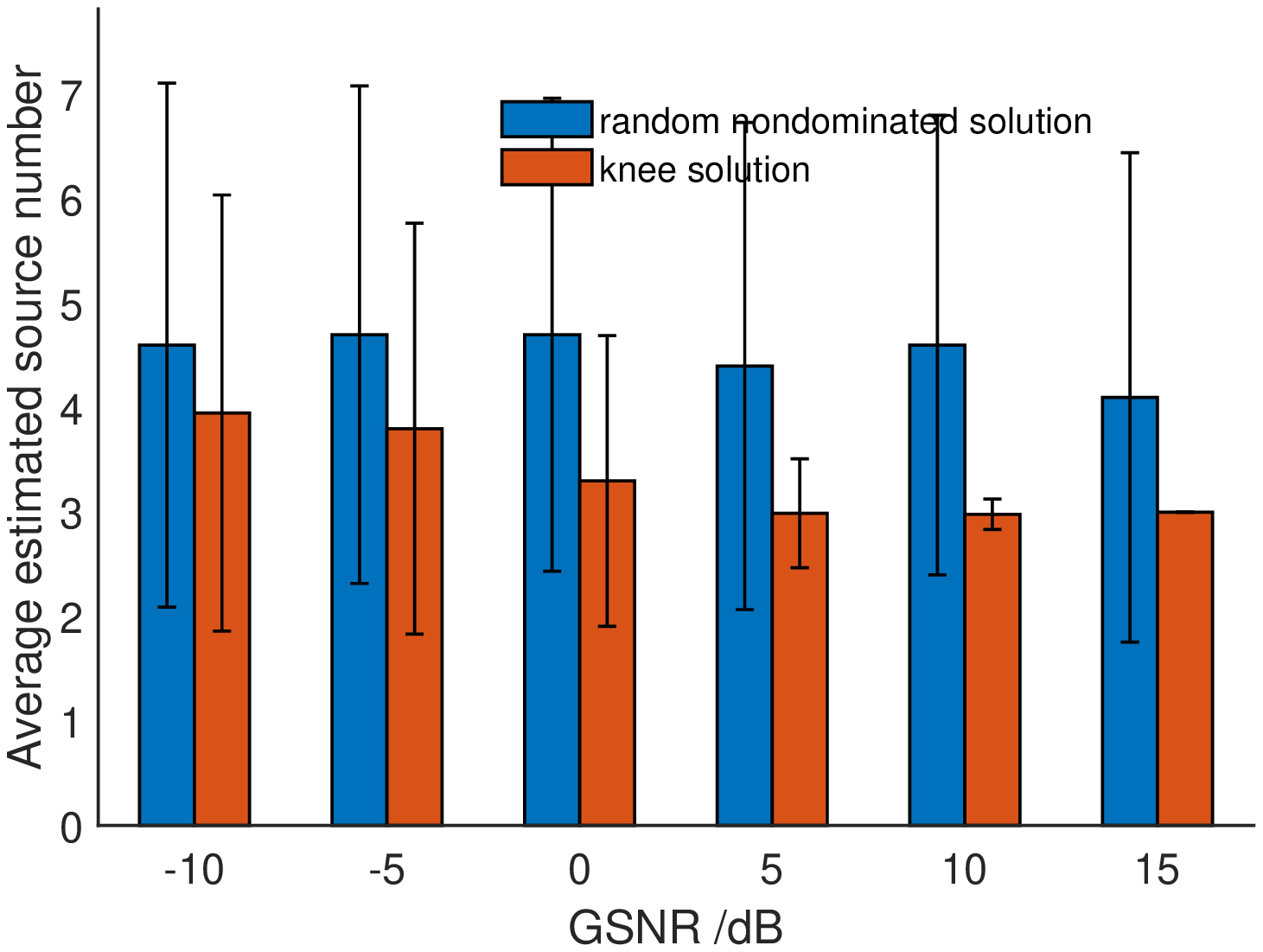}}
	\caption{Results of DOA estimate corresponding to the random nondominated solution and the knee solution under various $\text{GSNR}$s in S$\alpha$S noise ($\alpha=1.4$).}
	\label{fig-reKnee}
\end{figure}

To further exploit the impact of the knee solution on the accuracy of DOA estimate, the performance of MoBEA with the knee solution and a randomly selected nondominated solution entering the off-grid level optimization are compared in S$\alpha$S noise environment. The experimental parameters are the same as those in Fig. \ref{fig-PF} except that the GSNR varies from $-10\sim15$dB. Fig. \ref{fig-reKnee} exhibits the RMSE and average estimated source number versus $\text{GSNR}$. It can be seen that the version with the knee solution achieves a smaller DOAs error compared to the random nondominated solution case, thanks to its more precise estimate of the source number. This result indicates that the knee solution is more valuable than other nondominated solutions in terms of the source number and DOAs. 

\subsubsection{Efficacy of forward search}
 To validate the efficacy of forward search, the performance of MoBEA are compared to two versions, i.e., without the off-grid level optimization (denoted as the ``on-grid" version) and with Taylor expansion. In the last version, the manifold matrix is approximated using the first order Taylor approximation \cite{wu2018two}: $\At(\mta)=\At(\mta_0+\zta)\approx\At(\mta_0)+\At'(\mta_0)diag(\zta)$, where $\mta_0$ and $\zta$ are the initial grid and grid mismatch, respectively, $\At'(\mta_0)$ is the first derivative of $\At(\mta_0)$ with respective to $\mta_0$; similar to \cite{wu2018two}, $\zta$ can be obtained by letting the derivative of $V_{\text{CLF}}(\Yt,(\At(\mta_0)+\At'(\mta_0)diag(\zta))\St)$ with respect to $\zta$ be zero. 

\begin{figure}[] 
	\centering
	\subfigure[RMSE]{\includegraphics[width=0.45\textwidth]{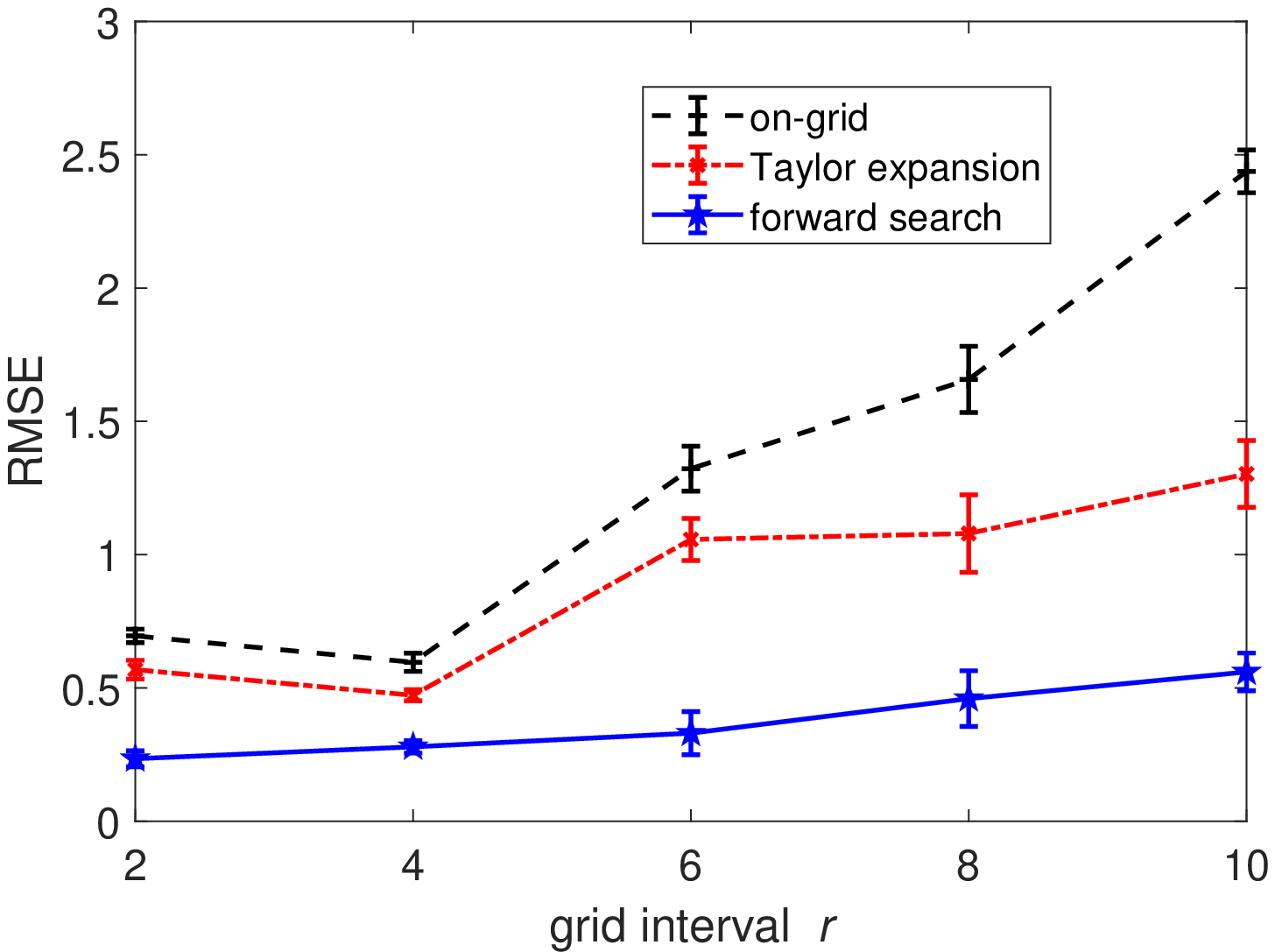}}
	\subfigure[Average estimated source number]{\includegraphics[width=0.45\textwidth]{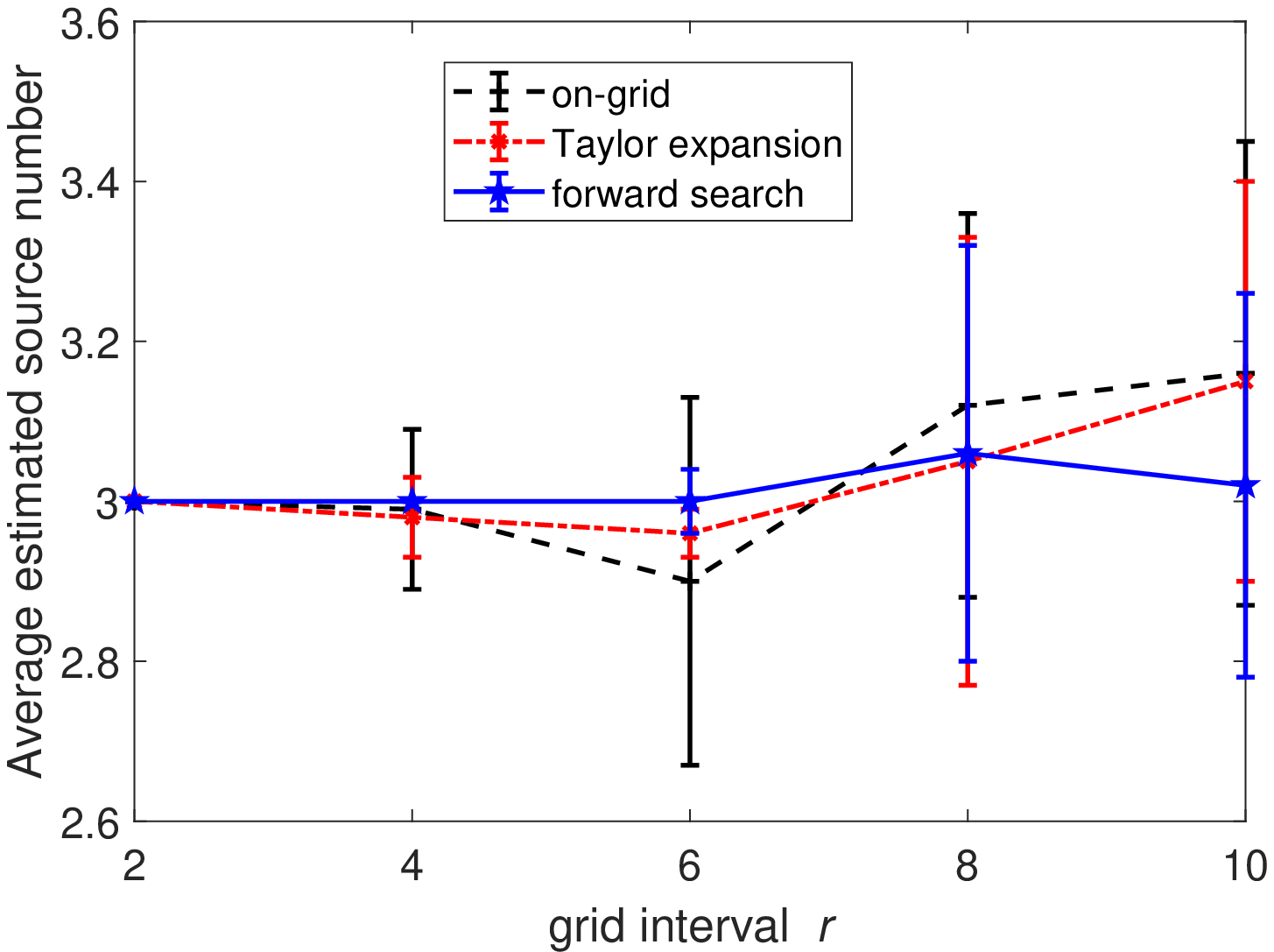}}
	\caption{Results of DOA estimate versus grid interval in GMM noise ($c_2=0.1$).}
	\label{fig-reoff_grid}
\end{figure}

In this simulation, the DOAs of two signals are from $1.6^{\circ}$ and $13.2^{\circ}$, respectively. The number of snapshots is $T=20$. Fig. \ref{fig-reoff_grid} shows the results of DOA estimate versus grid interval in GMM noise with $c_2=0.1$ and $\text{SNR}=10\text{dB}$. It is observed that the three strategies have similar results to the source number. The localization accuracy of the ``on-grid" version and ``Taylor expansion" version deteriorates with the coarser grid due to the ignorance of off-grid mismatch and the large modeling error caused by Taylor expansion, respectively. As expected, the MoBEA with forward search achieves the best performance with different grid intervals, which validates the forward search's superiority in evaluating the grid mismatch. 

\subsection{Comparison of MoBEA against State-of-the-Art Methods} \label{sec-exp-comparison}
In this subsection, we conduct experiments to compare MoBEA with the state-of-art methods, i.e., $l_p$-MUSIC \cite{Zeng2013}, MCC-MUSIC(\cite{wang2017a}), Bayes-optimal \cite{Dai2017Sparse}, and Fast-alternating \cite{Dai2017Sparse}.

\begin{figure}[t] 
	\centering
	\subfigure[RMSE]{\includegraphics[width=0.45\textwidth]{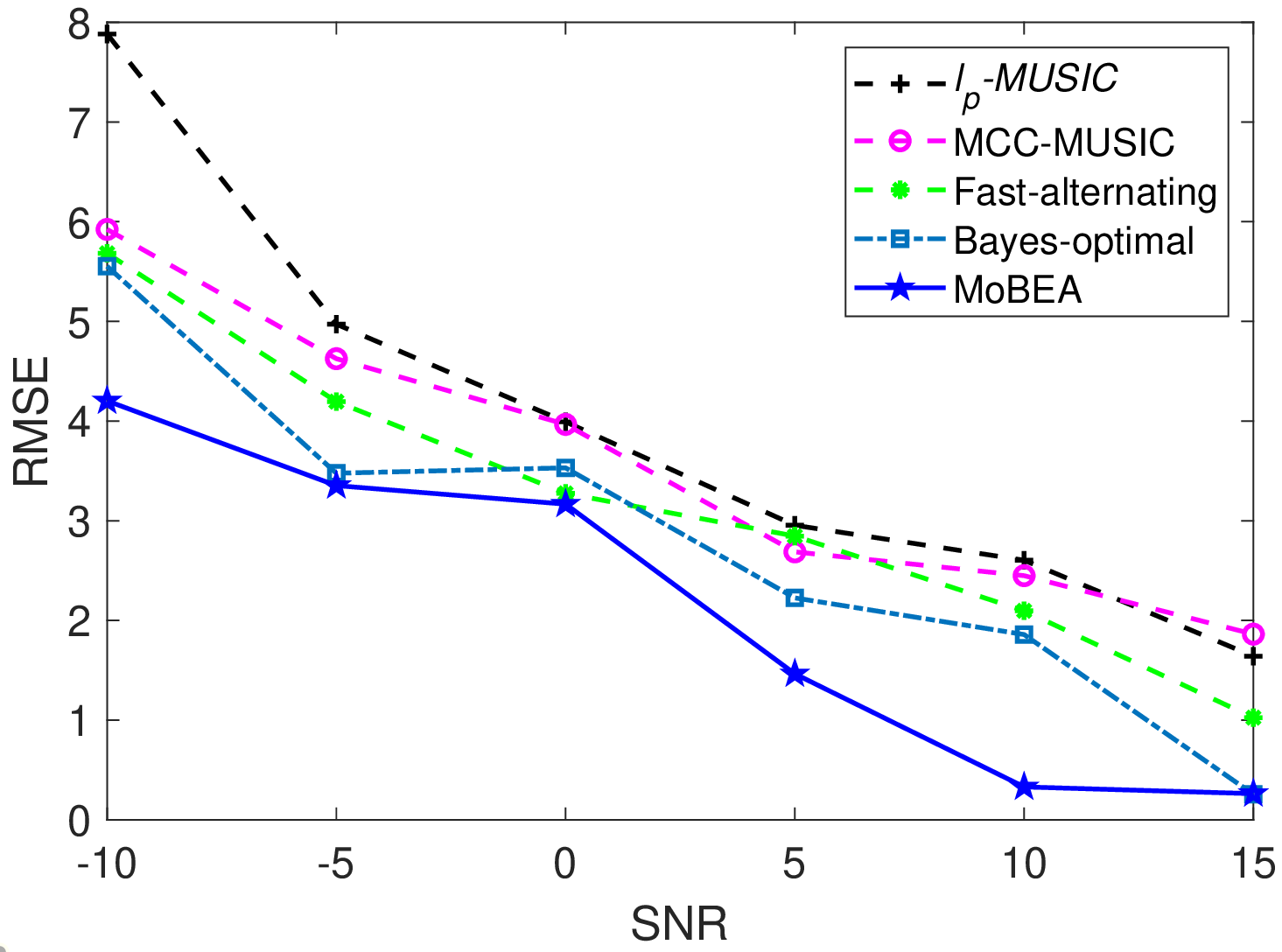}}
	\subfigure[Average estimated source number]{\includegraphics[width=0.45\textwidth]{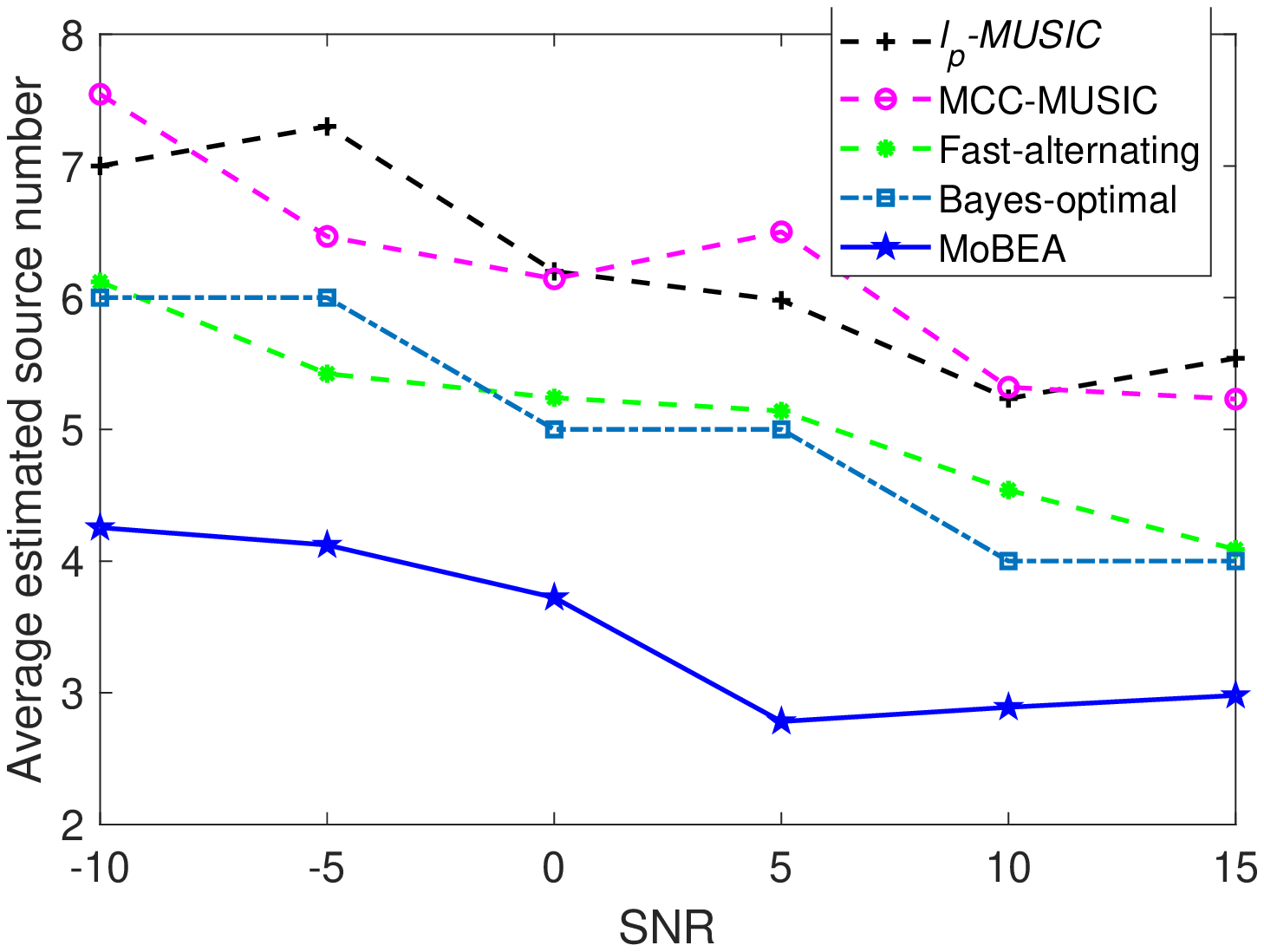}}
	\caption{Results of RMSE and average estimated source number versus SNR (dB) in GMM noise with $c=0.1$.}
	\label{fig-reSNR}
\end{figure}  

\subsubsection{Performance under various SNRs}
This subsection investigates algorithms' performance under different SNRs in GMM noise. Three uncorrelated sources from $-2.7^{\circ}$, $5.8^{\circ}$, and $20.2^{\circ}$ are considered. The number of snapshots is $20$. Fig. \ref{fig-reSNR} shows the RMSE and estimated source number in GMM noise versus SNR with $c_2=0.1$. It can be seen that the performance of all algorithms degrades obviously. $l_p$-MUSIC and MCC-MUSIC perform the worst under lower SNRs, which may be due to their high sensitivity to lower SNR and imprecise information of source number. The two Bayesian methods behave better in most cases, but they are inferior to MoBEA in localization accuracy. The reason is MoBEA estimating the source number more precisely, which avoids energy leakage on spurious DOAs. 

\begin{figure}[] 
	\centering
	\subfigure[RMSE]{\includegraphics[width=0.45\textwidth]{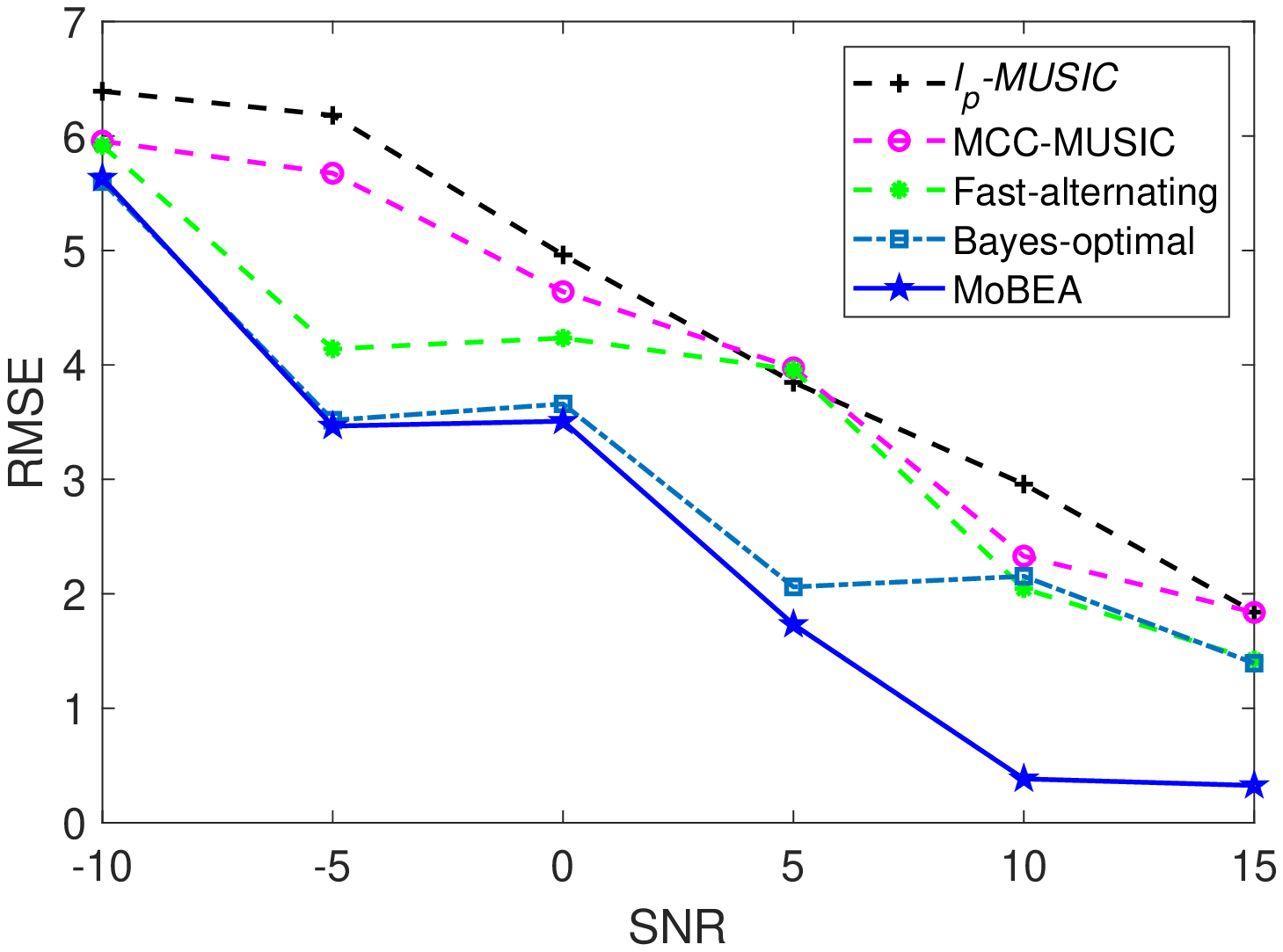}}
	\subfigure[Average estimated source number]{\includegraphics[width=0.45\textwidth]{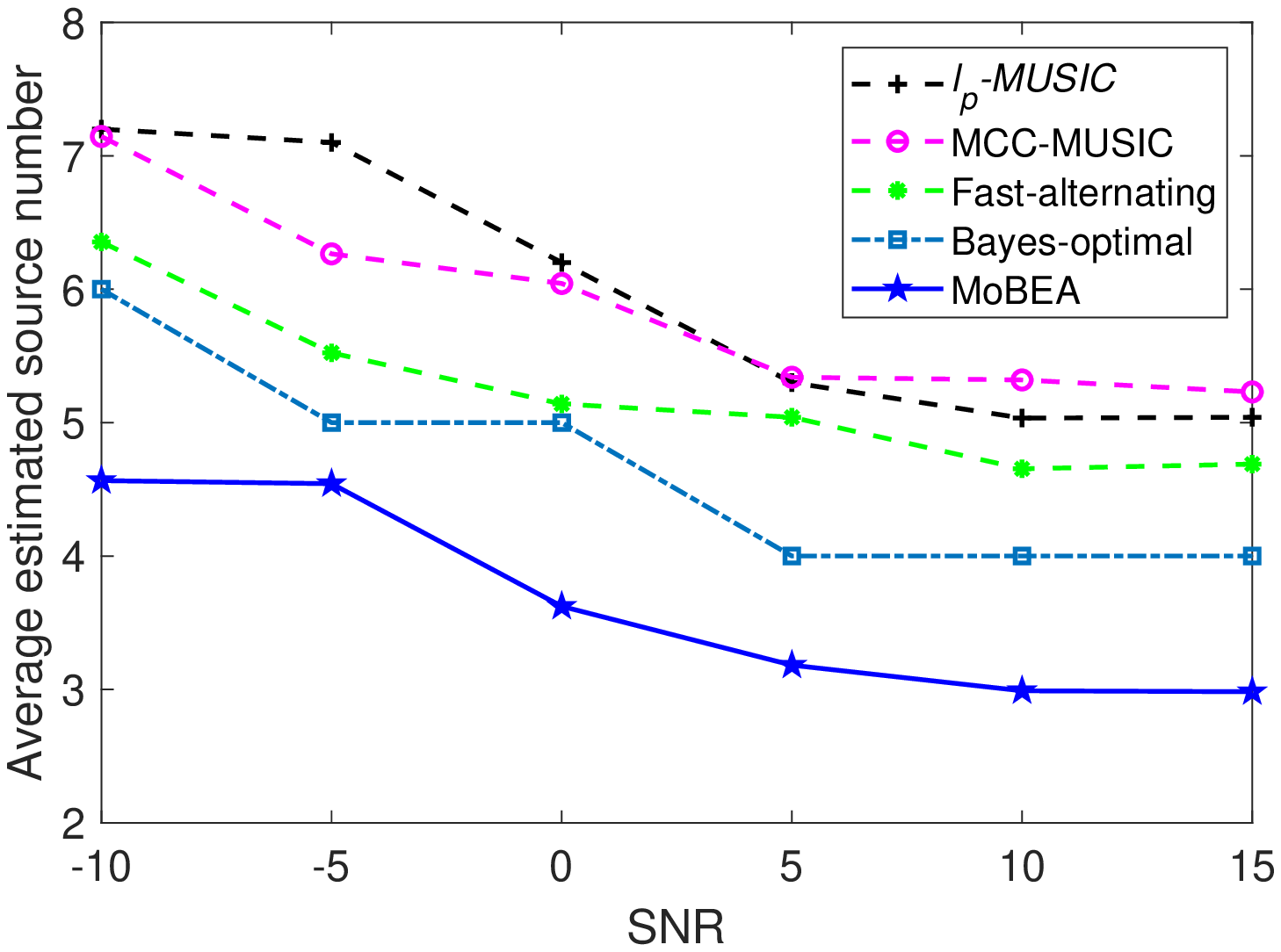}}
	\caption{Results of RMSE and average estimated source number versus SNR (dB) in S$\alpha$S noise with $\alpha=1.4$.}
	\label{fig-reSNR2}
\end{figure}

Fig. \ref{fig-reSNR2} reports the performance in S$\alpha$S noise versus GSNR with $\alpha=1.4$. The two MUSIC-based methods still behave badly in impulsive noise. Compared to the SBL-based methods, MoBEA achieves higher or comparable DOA estimation accuracy but always predicts the source number more accurately. 

\subsubsection{Performance under various grid intervals} 
In this simulation, the performance of MoBEA under various grid intervals is investigated. Three uncorrelated sources from $-19.7^{\circ}$, $6.8^{\circ}$, and $32.7^{\circ}$ are considered. The number of snapshots and SNR/GSNR are $20$ and $10\text{dB}$, respectively. The grid interval  varies from $2^{\circ}$ to $10^{\circ}$. 

\begin{figure}[t] 
	\centering
	\subfigure[RMSE]{\includegraphics[width=0.45\textwidth]{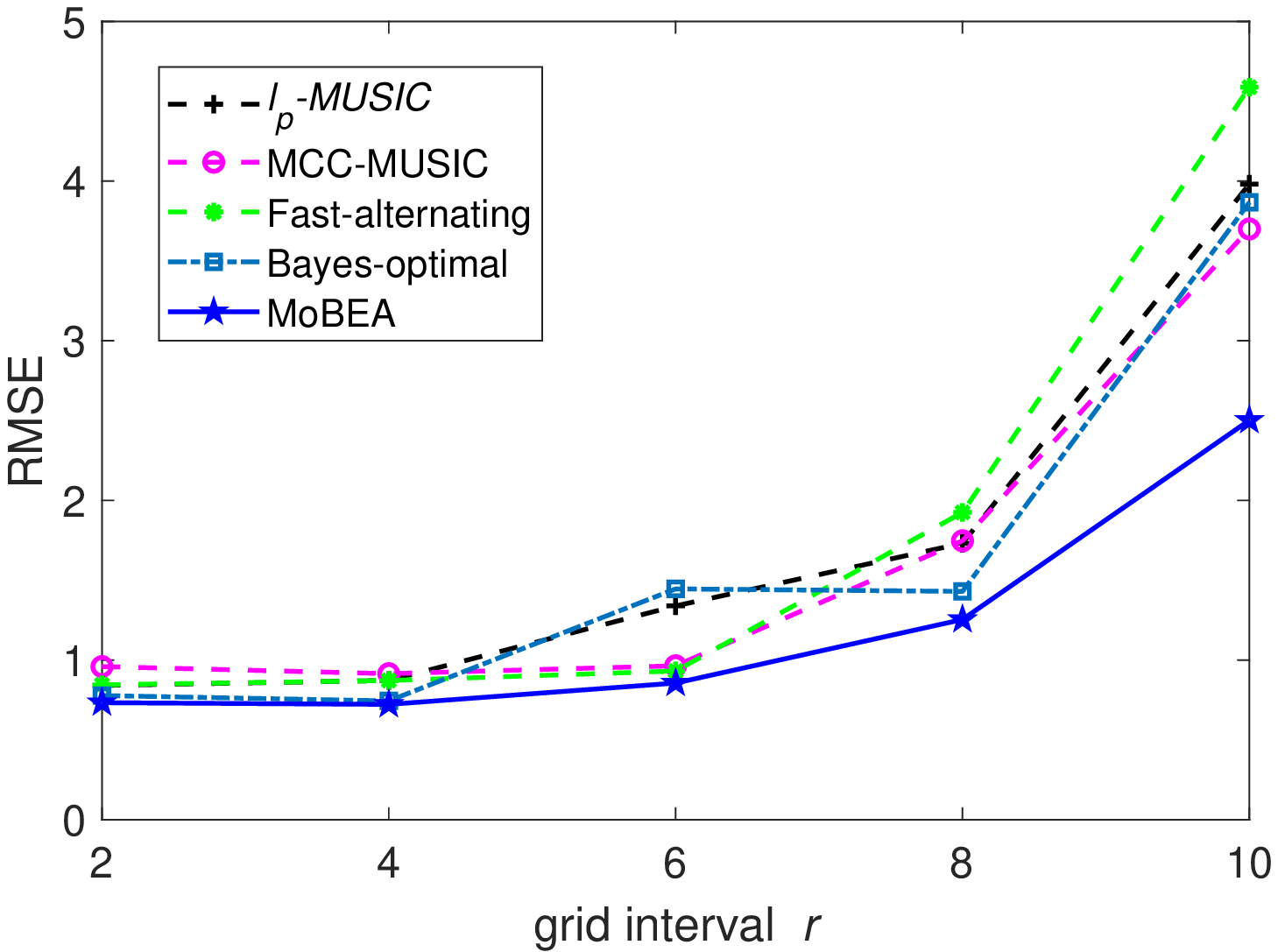}}
	\subfigure[Average estimated source number]{\includegraphics[width=0.45\textwidth]{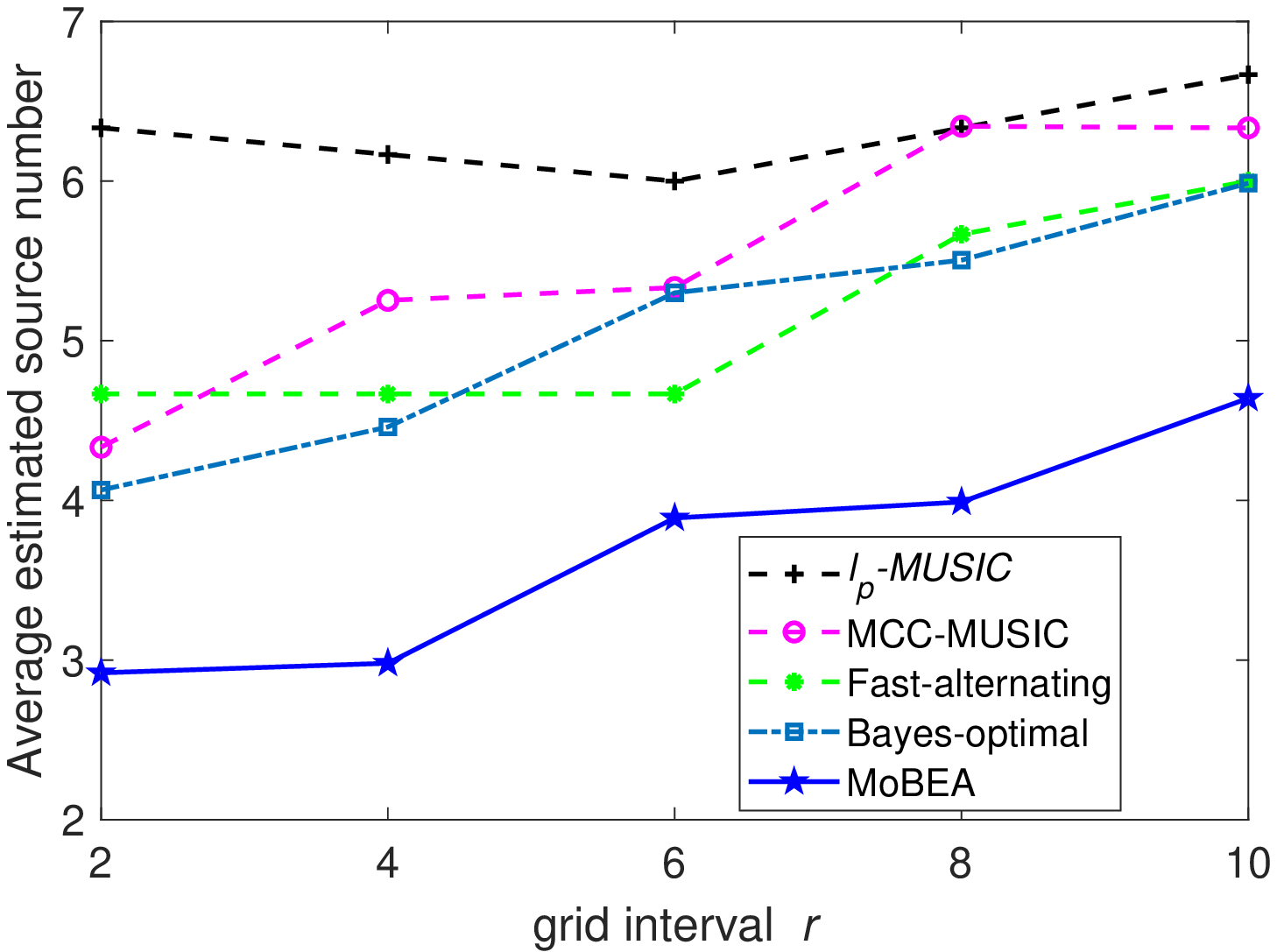}}
	\caption{Results of DOA estimate versus grid interval in S$\alpha$S noise with $\alpha=1.4$.}
	\label{fig-sas-reInterval}
\end{figure}

Fig. \ref{fig-sas-reInterval} exhibits the RMSE and average estimated source number versus grid interval in S$\alpha$S noise ($\alpha=1.4$). MoBEA achieves comparable or smaller RMSE results than other algorithms under lower SNRs and performs much better with higher SNRs. The enhancement of MoBEA under lower SNRs is non-significant, probably because the localization accuracy is severely affected by the ignorance of the noise variance in the proposed model. But it is worth noting that, MoBEA shows an absolute predominance in predicting the source number for all grid interval cases. 

\subsubsection{Performance under various angular separations}
\begin{figure}[] 
	\centering
	\subfigure[RMSE]{\includegraphics[width=0.45\textwidth]{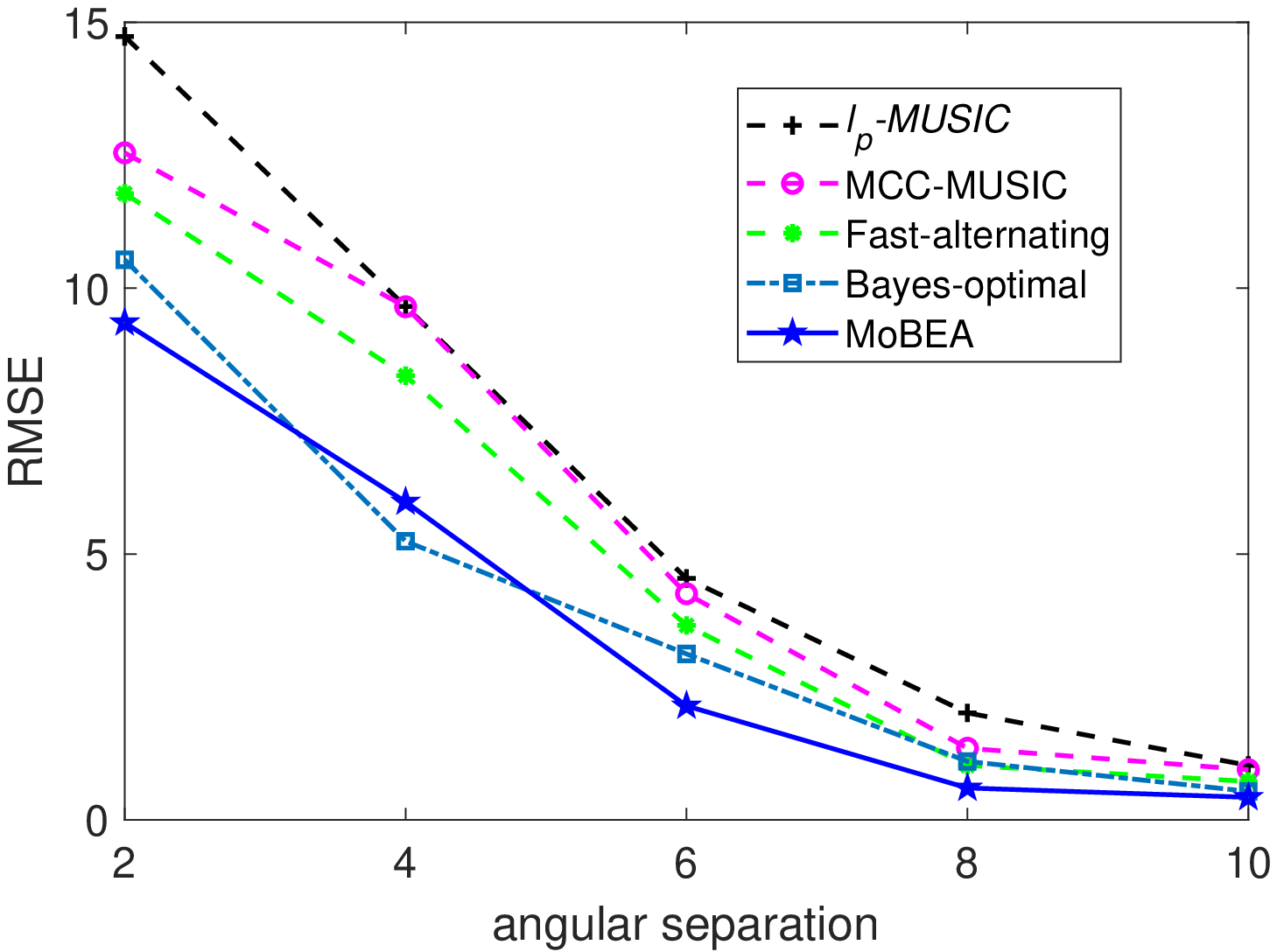}}
	\subfigure[Average estimated source number]{\includegraphics[width=0.45\textwidth]{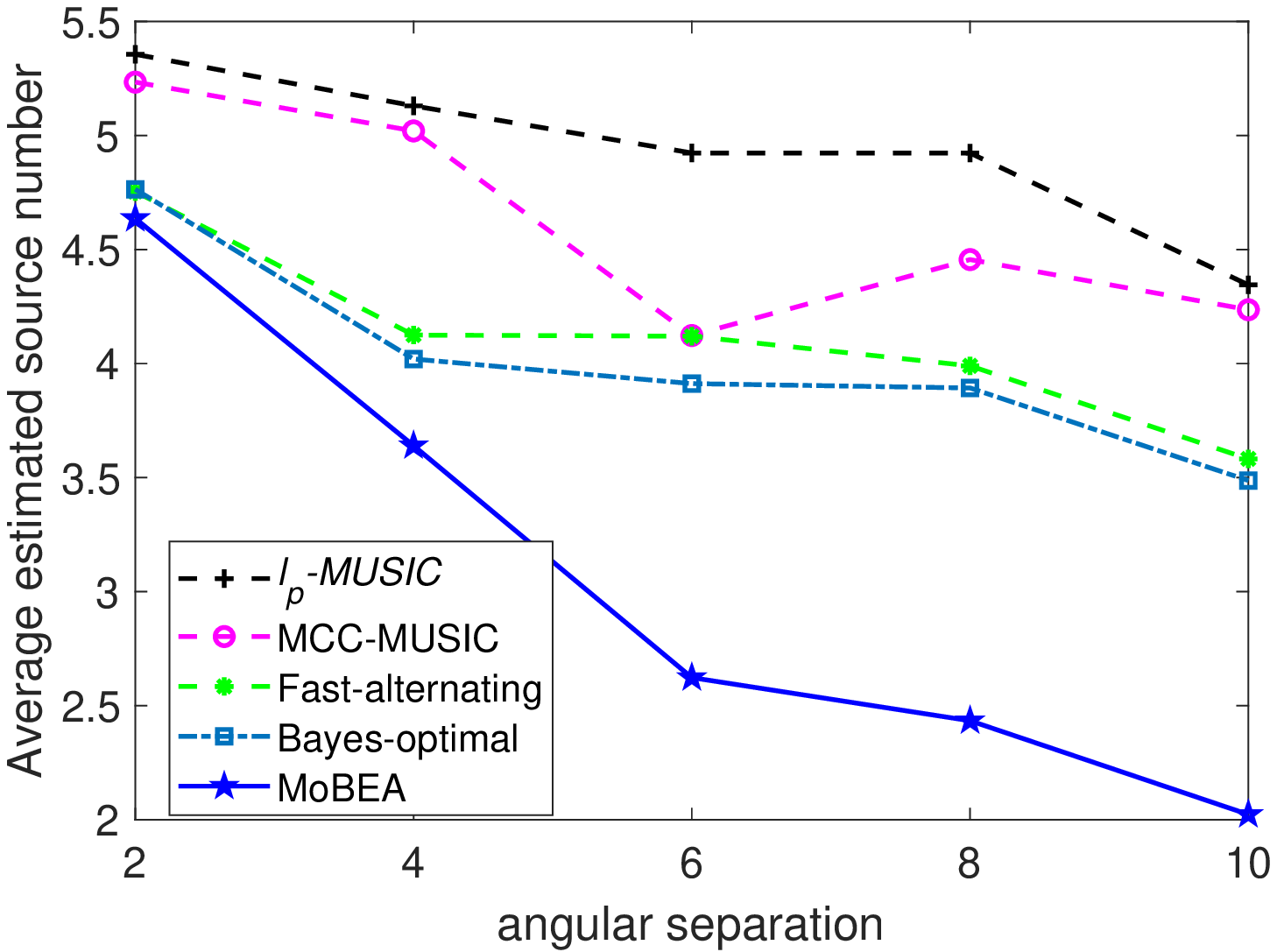}}
	\caption{The results of DOA estimate versus angular separation in GMM noise with $c_2=0.1$.}
	\label{fig-GMM-rereseparation}
\end{figure}

This simulation examines algorithms' ability to identify two closely-located sources. We consider two sources. The first is from $-10.8^{\circ}$ and the second varies from $-8.8^{\circ}$ to $-0.8^{\circ}$. The number of snapshots and $\text{SNR}$ are set to be $T=30$ and $10\text{dB}$, respectively. Fig. \ref{fig-GMM-rereseparation} displays the performance versus angular separations increasing from $2^{\circ}$ to $10^{\circ}$ in GMM noise. The RMSE values of MoBEA retain the lowest in most cases. As the two sources separate from each other, MoBEA predicts the source number more precisely, which performs remarkably better than other algorithms. These results indicate that MoBEA has higher localization resolution when the sources are closely located. 

\subsubsection{Performance under various numbers of snapshots}
In this simulation, the off-grid DOA estimation performance in S$\alpha$S noise with the number of snapshots at $\alpha=1.4$ and $\text{GSNR}=10\text{dB}$ is studied. Three sources are randomly chosen from $-2.7^{\circ}$, $5.7^{\circ}$ and $20.2^{\circ}$ impinging on the uniform linear array.

\begin{figure}[] 
	\centering
	\subfigure[RMSE]{\includegraphics[width=0.45\textwidth]{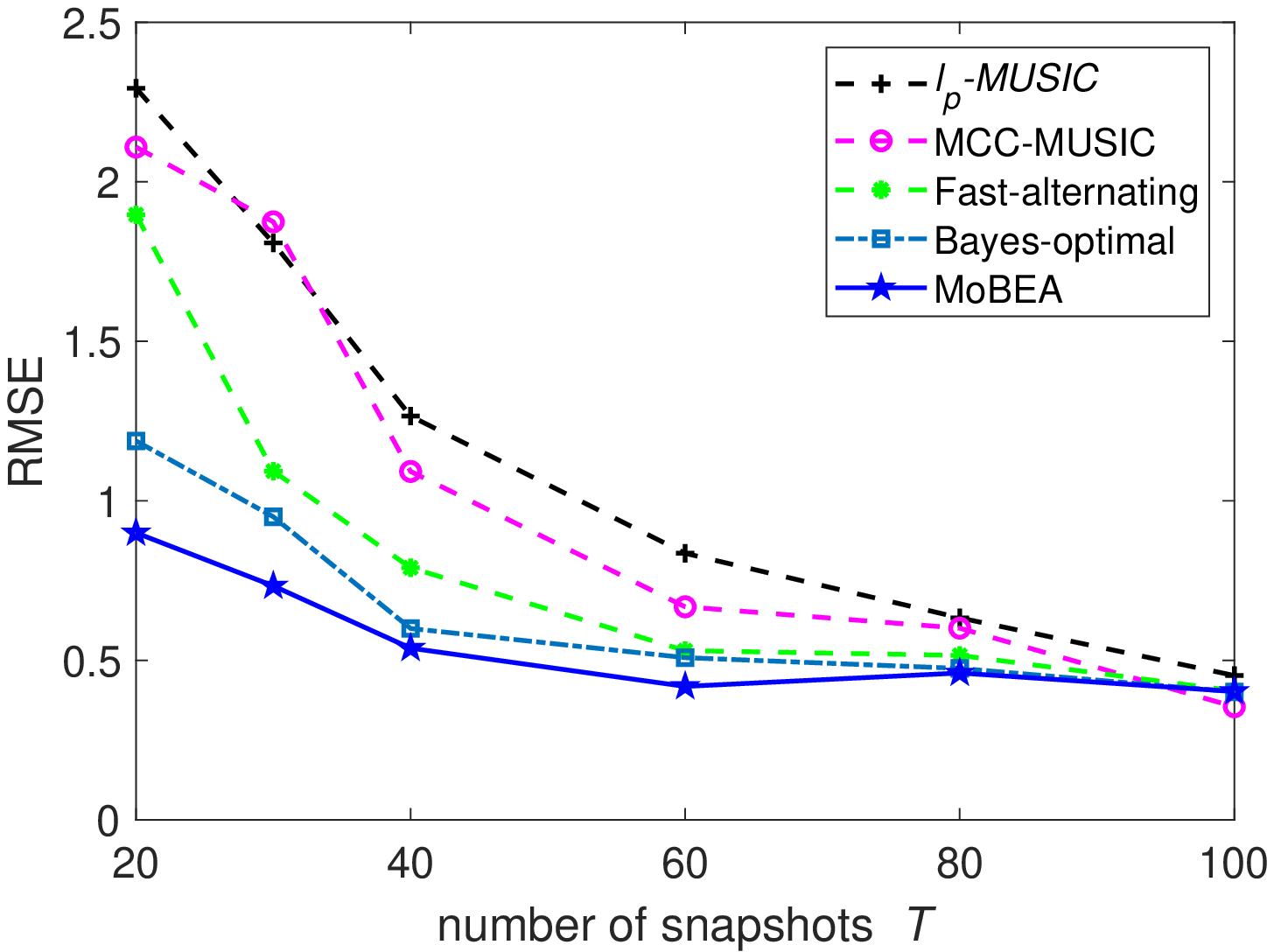}}
	\subfigure[Average estimated source number]{\includegraphics[width=0.45\textwidth]{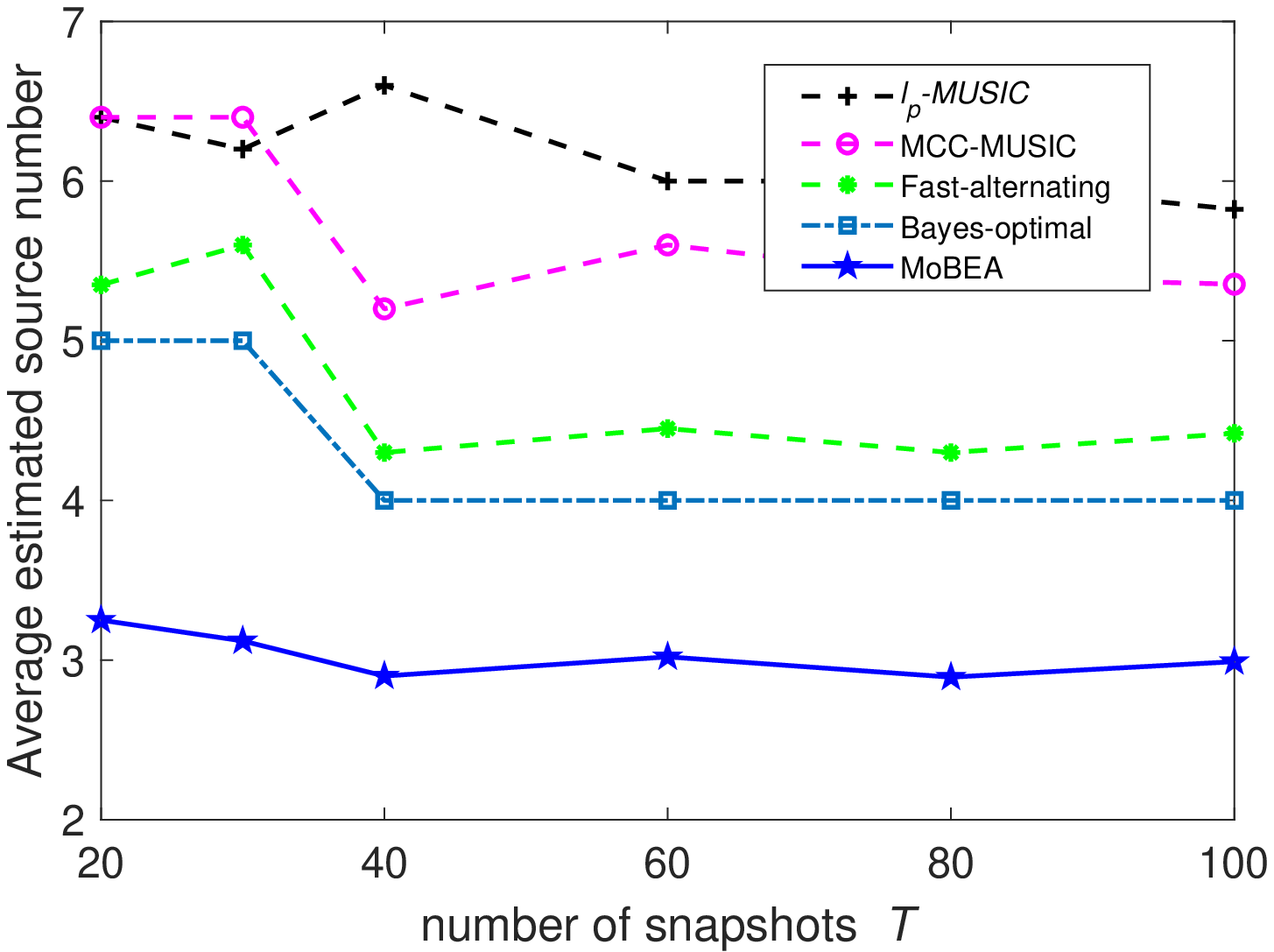}}
	\caption{The results of DOA estimate versus number of snapshots in S$\alpha$S noise with $\alpha=1.4$.}
	\label{fig-sas-reT}
\end{figure}

Fig. \ref{fig-sas-reT} shows the RMSE and estimated source number in S$\alpha$S noise versus the number of snapshots. It can be observed that the $l_p$-MUSIC and MCC-MUSIC algorithms gain the worst performance under fewer data samples. Fast-alternating and Bayes-optimal methods obtain great enhancement in RMSE with fewer snapshots. As expected, MoBEA achieves the highest localization accuracy due to the accurate identification of the source number. 

\subsubsection{Performance on running time}
\begin{figure}[t]
	\centering
	\includegraphics[width=0.5\columnwidth]{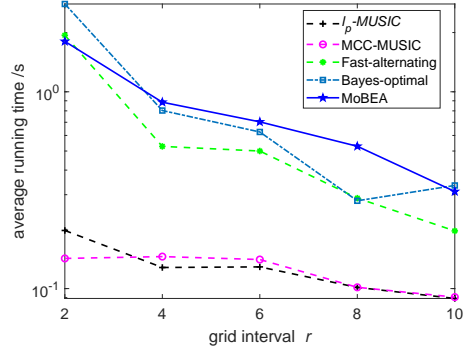}
	\caption{The average running time of DOA estimate versus grid interval in S$\alpha$S noise with $\alpha=1.4$.}
	\label{fig-running-time}
\end{figure}

Section \ref{sec-complexity} demonstrates that the number of grid points $N$ has a remarkable impact on the computational complexity of MoBEA. To further quantify the impact, this simulation provides the average running time of MoBEA versus the grid interval compared to other algorithms, as shown in Fig. \ref{fig-running-time}. The experimental settings are the same as those in Fig. \ref{fig-sas-reInterval}. It can be observed that the computational time of all algorithms shows a decreasing trend with increasing grid intervals (i.e., fewer grid points). The $l_p$-MUSIC and MCC-MUSIC run much faster than other algorithms. MoBEA and Bayes-based algorithms are slower because they need to execute the time-consuming inversions. If there is a high demand for the faster running speed of MoBEA in practical applications, the decoding is suggested to be executed in parallel. Despite this, MoBEA achieves great improvements in RMSE and estimated source number over the comparing algorithms in most scenarios.    

\section{Conclusion}
This paper has proposed the MoBEA for DOA estimation. MoBEA involves two innovations. The first is the multiobjective DOA estimation model, in which the source number and a robust correntropy-based fitting error function are taken as two objectives. Unlike existing DOA estimation models, the proposed model automatically identifies the source number together with DOA estimation. Besides, the original $l_0$-norm penalty is used to capture the sparsity of signal-of-interest, which avoids relaxing sparse-inducing penalties and enables source number identification more accurately. The second innovation is the multiobjective bilevel evolutionary DOA estimation algorithm for solving the proposed model. The on-grid level is for joint source number identification and sparse recovery, while the off-grid level works on grid refinement via the proposed forward search strategy. This strategy avoids linear approximation and enhances the localization accuracy. Thanks to the population-based search of the proposed algorithm, the solutions own different source numbers communicate to each other during the evolutionary search, which provides diverse search pathways to the optima. Experimental results have demonstrated the effectiveness and efficiency of the proposed MoBEA in identifying the source number and DOAs in impulsive noise. 

In the future, parallel processing will be considered to accelerate our algorithm. A future study on reducing the computational complexity of MoBEA will be also explored.

\bibliography{Ref}

\end{document}